%% file: neurips_2026.tex
\documentclass{article}

\PassOptionsToPackage{numbers,compress}{natbib}
\usepackage[main, final]{neurips_2026}

\usepackage[table]{xcolor}

\definecolor{myblue}{RGB}{230,240,255}
\definecolor{mypurple}{RGB}{240,230,255}
\usepackage[utf8]{inputenc}
\usepackage[T1]{fontenc}
\usepackage{hyperref}
\usepackage{url}
\usepackage{booktabs}
\usepackage{amsfonts}
\usepackage{nicefrac}
\usepackage{microtype}
\usepackage[table]{xcolor}
\usepackage{graphicx}
\usepackage{subcaption}
\usepackage{tabularx}
\usepackage{amsmath}
\usepackage{array}
\usepackage{wrapfig}
\usepackage{enumitem}
\usepackage{amsthm}
\usepackage{multirow}
\usepackage{caption}
\usepackage{algorithm}
\usepackage{algpseudocode}
\newcommand{\flowarrow}{\ensuremath{\rightarrow}}

\title{LoopWeaver: Weaving Feedback Loops into Hierarchical Generation under Constraints}

\author{%
Yifan Zhu\textsuperscript{1}\hspace{0.5em}
Guanting Chen\textsuperscript{1}\hspace{0.5em}
Bing Wei\textsuperscript{3}\hspace{0.5em}
Haoran Luo\textsuperscript{2}
\thanks{Corresponding author}
\\[0.5em]
\textsuperscript{1}Beijing University of Posts and Telecommunications\\
\textsuperscript{2}Nanyang Technological University, Singapore\\
\textsuperscript{3}Hainan University\\[0.3em]
\texttt{\{yifan\_zhu,cgt\}@bupt.edu.cn, haoran.luo@ntu.edu.sg}
}

\begin{document}

\maketitle

\begin{abstract}
    Large Language Models face a multi-objective optimization problem in constrained long-text generation, balancing global structure, local coherence, and constraint satisfaction. We propose LoopWeaver, a hierarchical workflow with two sequential offline preference-optimization stages. Constraint-aware planning produces structured plans, feasibility filtering removes invalid plans, and a shared deterministic reward constructs preference pairs for plan-level and generation-level DPO. At inference time, LoopWeaver remains feed-forward: it generates and revises a complete plan, filters the revised plan, and then generates segments conditioned on the current sub-plan and running narrative. We evaluate the method  on four LongGenBench-style structured scenarios across multiple backbones. Results show improvements in both constraint accuracy and text quality across all four open-source backbones. Our software and data are publicly available at \url{https://github.com/cgtaa-cmyk/LoopWeaver}.
\end{abstract}

\section{Introduction}
\label{Introduction}

\begin{wrapfigure}{r}{0.55\textwidth}
\vspace{-4.5mm}
\centering
\includegraphics[width=7.5cm]{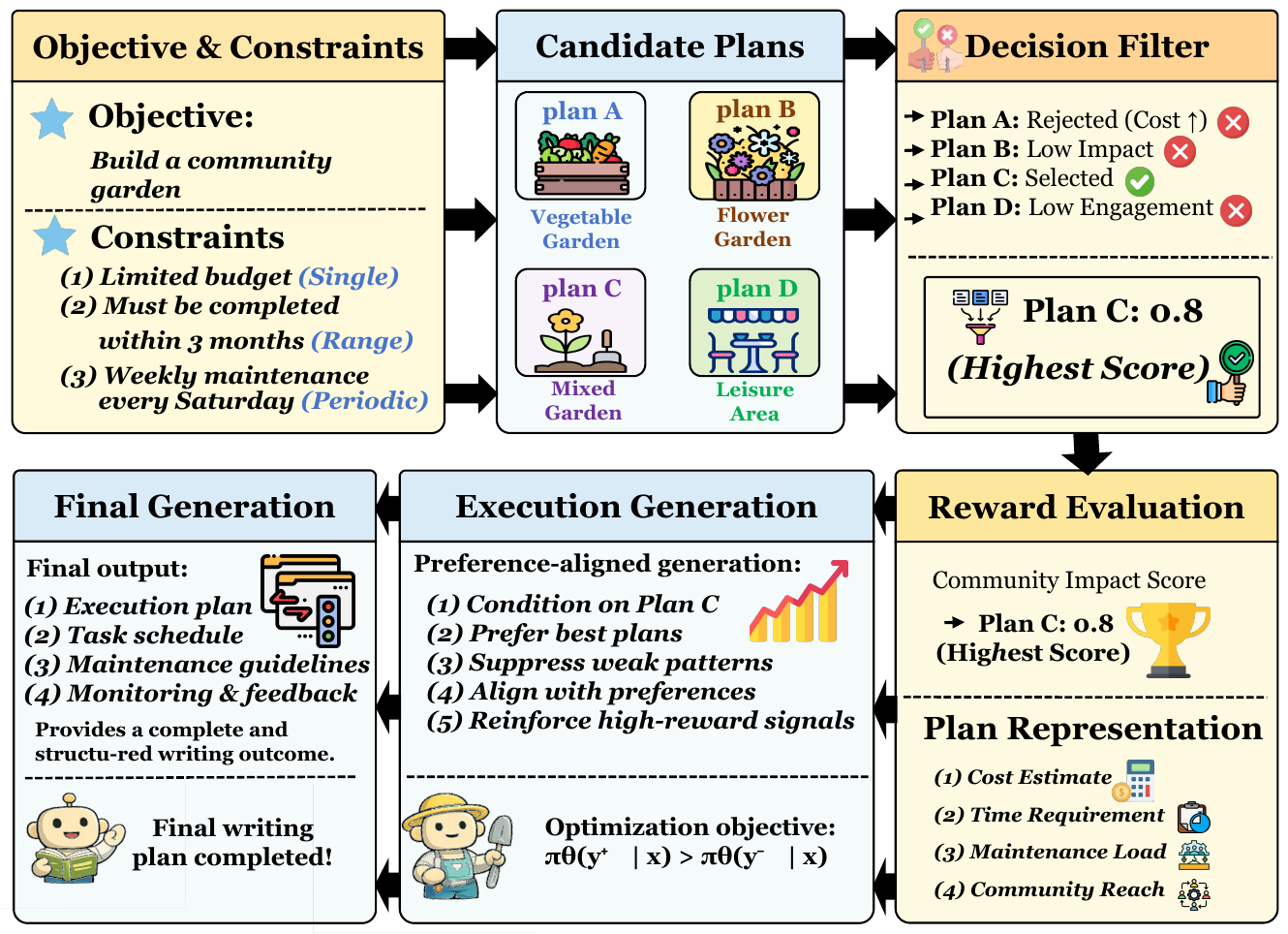}
\vspace{-0.5mm}
\caption{Constrained long-form generation in LoopWeaver with planning, generation, and verification.}
\label{fig_1}
\vspace{-1mm}
\end{wrapfigure}

Large language models~\citep{zhao2023survey} excel at short-form text generation~\citep{radford2019language, brown2020language, achiam2023gpt}, but struggle with \emph{constrained long-form generation}~\citep{bai2024longwriter, xia2025storywriter, xi2025omnithink, EIPE-text, lee2025navigating, IntegratingPlanning, wu2025shifting, xu2025topic, HelloBench}, which requires outputs of thousands or even ten thousand tokens that meet diverse user-specified constraints (e.g., length, structure, style, or factual requirements) while preserving global coherence~\citep{zhou2023controlled, chen2022controllable, EvalConstrained}. This entails jointly optimizing global structure, local coherence, and constraint satisfaction, three tightly coupled objectives that are hard to balance~\citep{cho2019towards, CTGSurvey, HiCaM}. As illustrated in Figure~\ref{fig_1}, the task unfolds as a multi-stage process interleaving planning, generation, and constraint verification~\citep{wang2025self, han2025stitch, SuperWriter, LongGenBench}.

A promising direction is to cast constrained long-form writing as a workflow, where planning, generation, verification, and revision are first-class stages connected by structured information flow~\citep{wang2025self, SuperWriter, WorkflowLLM, Flow}. This view aligns with the rise of agentic LLM systems, coding agents such as Claude Code, and frameworks such as OpenHands, showing how workflow-style decomposition extends LLM capabilities on complex, multi-step tasks~\citep{ChainOfAgents, LiuLXZP25, YuanSCTLY25, abs-2505-21116, CreAgentive, luo2025graphr1}. Early efforts follow this trend by decomposing long-form writing into hierarchical or staged pipelines that interleave outlining, drafting, and refinement~\citep{WanMH0GC25, xia2025storywriter, EIPE-text, IntegratingPlanning}. Together, these developments suggest that explicit workflow-level modeling rather than monolithic decoding or rigid pipelines alone is a natural fit for the coupled optimization such tasks require.

Existing approaches largely fall into two paradigms, neither fully resolving the coupled optimization above. \textbf{Data-centric methods} use data augmentation or preference optimization to produce longer, more constraint-aware outputs~\citep{bai2024longwriter, APO, DPO_and_RLHF, rafailov2023direct, wu2025longwriter, ping-etal-2025-longdpo, salemi2025reasoning, deng2022rlprompt, wang2024esrl}, but encode constraints implicitly and cannot verify or repair violations at inference, letting errors accumulate~\citep{liu2024lost, bastan2023neurostructural, PynadathZ25}. \textbf{Planning-based methods} decompose generation into hierarchical outlines or multi-agent pipelines~\citep{xia2025storywriter, YaoYZS00N23, ShinnCGNY23, EIPE-text, WanMH0GC25, roy-etal-2024-contregen, WikiWriter, RAPID}, yet planning, generation, and verification stay loosely coupled: planning-time constraints rarely reach segment-level decoding, and later errors seldom revise earlier plans~\citep{ZhangHLCHNHDFL25, BeyondICL, Suri, Collie}. Neither fully and jointly optimizes the three objectives, motivating a unified offline preference-optimization formulation.

\begin{figure*}[t]
\centering
\includegraphics[width=1.0\linewidth]{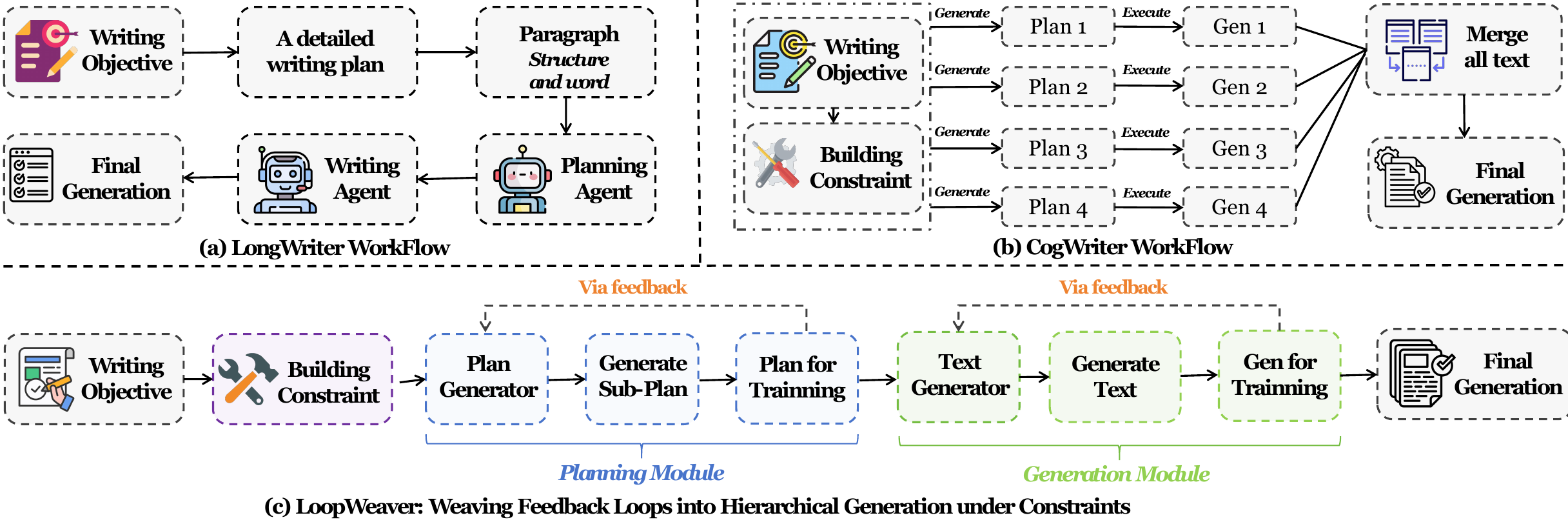}
\caption{Framework comparison of LoopWeaver, LongWriter, and CogWriter. LoopWeaver couples plan-level and generation-level preference optimization during offline training, while inference proceeds through complete-plan revision, filtering, and segment generation.}
\label{F2}
\end{figure*}

However, when transplanted to constrained long-form writing, existing workflow-style systems still expose three recurring limitations: \textbf{(i) No constraint-aware modeling.} Generic workflows orchestrate sub-tasks without propagating user constraints through outlining, drafting, and assembly~\citep{BonlarronRMR25, AFlow, ADAS, dspy}. \textbf{(ii) Limited adaptivity.} Most pipelines use static flows or costly post-hoc refinement, still unable to fully adapt to changing constraints, length, or intermediate signals~\citep{AdaptFlow, EvoFlow, mass, maestro, gepa}. \textbf{(iii) Loose feedback coupling.} Feedback in Claude Code and OpenHands is local and post hoc, with no closed loop letting segment-level signals revise the global plan or vice versa~\citep{GulA24, ZhangHLCHNHDFL25, AGP, agentdropout, flowreasoner, robustflow}.

To address these limitations, we propose \textbf{LoopWeaver}, a hierarchical workflow whose two preference-optimization stages are coupled during offline training~\citep{maas, ayc, a2flow}. \emph{Constraint-aware hierarchical planning} decomposes user goals into a complete structured plan with constraints propagated to its generation units~\citep{IntegratingPlanning, EIPE-text, kou2026pu}. \emph{Feasibility filtering} removes plans that fail prompted checks for relevance, completeness, coherence, specificity, efficiency, and consistency~\citep{WanMH0GC25, mermaidflow, vflow}. Finally, \emph{reward-based preference optimization} uses a shared deterministic reward to construct plan-level and generation-level preference pairs~\citep{WangHLWLHT25, rafailov2023direct, ping-etal-2025-longdpo}. The resulting policies are executed feed-forward at inference time; generated segments are not returned to the planner.

We evaluate LoopWeaver on four LongGenBench-style structured constrained-generation scenarios across multiple backbone models, including the Qwen2.5~\citep{Qwen2.5} series and LLaMA3.1-8B-Instruct~\citep{Llama3}, and compare it with both proprietary and open-source baselines such as GPT-4o-mini~\citep{achiam2023gpt}, as well as representative workflow-based and data-centric methods including CogWriter~\citep{WanMH0GC25} and LongWriter~\citep{bai2024longwriter, wu2025longwriter}. Across metrics covering constraint satisfaction and narrative-level quality~\citep{LongGenBench, HelloBench, EvalConstrained}, the revised audit reports where LoopWeaver improves over baselines and where it does not, rather than assuming a uniform gain across backbones; these results do not constitute evidence of unrestricted cross-domain generalization. These findings highlight the benefits of coordinated planning and generation.

\section{Related Work}
\subsection{Constrained Long-Form Text Generation.} 
Constrained long-form text generation aims to produce coherent long texts under task-specific requirements such as content, style, length, and structural constraints~\citep{HiCaM, EvalConstrained, xu2025topic, luo2025kbqao1}. Prior work mainly falls into two paradigms~\citep{CTGSurvey}: controllable generation via explicit constraint modeling, which uses instruction tuning, decoding-time control, or RL-based prompt optimization to enforce user-specified properties~\citep{BeyondICL, zhou2023controlled, deng2022rlprompt, Collie, Suri}, and planning-based generation, where hierarchical outlines or retrieval-augmented contexts improve global coherence and structural consistency~\citep{WanMH0GC25, WikiWriter, EIPE-text}. A complementary line scales long-form generation through large-context training and preference optimization~\citep{bai2024longwriter, wu2025longwriter, ping-etal-2025-longdpo}. Despite this progress, existing methods typically adopt loosely coupled pipelines~\citep{PynadathZ25, RAPID, kou2026fedharmony}, limiting the integration of constraint signals and feedback during generation, which often leads to compounding errors and degraded coherence in long outputs~\citep{liu2024lost, LongGenBench, HelloBench}.

\subsection{Agentic Workflows}
Recent work abstracts agentic workflows as executable computation graphs, dividing existing methods into static and dynamic approaches~\citep{WorkflowLLM, EvoFlow, Flow}. Static methods commit to reusable templates offline via MCTS-based operator search~\citep{AFlow}, code-space refinement~\citep{ADAS}, abstract operator libraries~\citep{a2flow}, prompt compilation or reflective evolution within fixed scaffolds~\citep{dspy, gepa}, joint structure-configuration updates~\citep{mass, maestro}, and validation-embedded search~\citep{mermaidflow, vflow}; related lines explore tool-orchestrated reasoning~\citep{YaoYZS00N23, ShinnCGNY23} and chain-of-agents pipelines~\citep{ChainOfAgents, LiuLXZP25}. Dynamic methods defer structural decisions to inference time, either pruning redundant edges from fixed hypergraphs~\citep{AGP, agentdropout} or synthesizing per-query workflows via autoregressive DAG generation~\citep{ayc}, agentic supernets~\citep{maas}, RL meta-agents~\citep{flowreasoner}, or preference learning~\citep{robustflow, AdaptFlow}. While these advances improve task generality and adaptivity, they primarily target reasoning, tool use, or code generation, rarely incorporating explicit constraint propagation or feedback-coupled coordination for constrained long-form text generation~\citep{YuanSCTLY25, abs-2505-21116, BonlarronRMR25}.

% Recent work increasingly frames complex generation as agentic or workflow-based processes, decomposing generation into coordinated stages or interacting agents~\citep{LiuLXZP25,YuanSCTLY25,abs-2505-21116}.
% Prior studies explore automated workflow construction via modular design, search, or evolutionary strategies, as well as explicit workflow orchestration for multi-step reasoning~\citep{AFlow,EvoFlow,Flow,WorkflowLLM}, while other approaches investigate multi-agent collaboration through end-to-end training or distillation~\citep{ChainOfAgents,CreAgentive}.
% In parallel, planning-based paradigms have been applied to long-form text generation via agent-based decomposition, hierarchical planning, or staged prompting to improve length and coherence~\citep{bai2024longwriter,IntegratingPlanning,HelloBench}.
% Despite their effectiveness, these methods largely emphasize workflow construction or decomposition and lack dynamic, adaptive, and constraint-aware end-to-end feedback-coupled optimization that jointly governs planning and generation.

\section{Preliminaries}
\label{sec:preliminaries}
This section introduces the reward-scoring and preference-optimization components used in LoopWeaver's two sequential offline DPO stages. % \textbf{(i) Constrained Long-Form Text Generation.} The task generates text satisfying constraints using controllable generation or fine-tuning~\citep{bastan2023neurostructural, chen2022controllable}, enabling control over structure, style, and content.

\textbf{(i) Deterministic Unit-Level Reward Evaluation.} We score each candidate sub-plan or generated segment under the shared rule-based reward. No additional rollouts occur for reward estimation.
\begin{equation}
r(z,e) = R(z,e),
\end{equation}
where $z$ is the candidate unit and $e=(x,C)$ is its prompt-and-constraint context.

\textbf{(ii) Reward Function $R(\cdot)$.}
$R(\cdot)$ is a \emph{deterministic, rule-based} scoring function. Given a candidate unit $s$ and context $e = (x, C)$ with $C = C_{\text{once}} \cup C_{\text{range}} \cup C_{\text{periodic}}$, it returns a scalar in $[0, 1]$:
\begin{equation}
R(s, e) = \min\!\big(1,\; r_{\text{constraint}}(s, C)
+ \alpha \cdot r_{\text{relevance}}(s, x)\big),\end{equation}
\begin{equation}
\label{eq:constraint-r}
r_{\text{constraint}}(s, C) = \frac{1}{|C|} \sum_{c \in C} \mathbf{1}\!\left[\rho(s,c) \geq \tau\right],
\end{equation}
where $x$ is the input prompt and $e=(x,C)$ is the prompt-and-constraint context.
The relevance term $r_{\text{relevance}}(s,x)$ measures prompt-keyword coverage with
$\alpha=0.5$.
\begin{equation}
\rho(s,c) =
\begin{cases}
1, & c \subseteq s,\\
\operatorname{SequenceMatcher.ratio}(s,c), & \text{otherwise},
\end{cases}
\qquad \tau=0.7.
\end{equation}
Thus, the reward measures constraint coverage and prompt relevance only; it is not a direct evaluator of tone, factuality, or document-level coherence. Candidate units are ranked by $R(s,e)$ to construct DPO preference pairs.

% \textbf{(iv) Direct Preference Optimization.} We use the standard DPO loss~\citep{rafailov2023direct},
% $\mathcal{L}_{\text{DPO}}(\theta) = -\mathbb{E}_{(z^+, z^-) \sim D}\!\left[\log \sigma\!\left(\beta \log \frac{\pi_\theta(z^+|x)\,\pi_{\text{ref}}(z^-|x)}{\pi_\theta(z^-|x)\,\pi_{\text{ref}}(z^+|x)}\right)\right]$,
% where preference pairs $(z^+, z^-) \sim D$ are ranked by deterministic reward $r(\cdot)$ and $\beta$ controls deviation from $\pi_{\text{ref}}$.

\textbf{(iii) Direct Preference Optimization.} 
DPO aligns preferences by optimizing the likelihood of higher- and lower-preference outputs \(z^+\) and \(z^-\) based on the deterministic reward, improving alignment.
\begin{equation}
\mathcal{L}_{\text{DPO}}(\theta)
= - \mathbb{E}_{(z^+, z^-) \sim D}
\left[\log \sigma\!\left(
\beta \log
\frac{
\pi_\theta(z^+ \mid x,C)\,\pi_{\text{ref}}(z^- \mid x,C)
}{
\pi_\theta(z^- \mid x,C)\,\pi_{\text{ref}}(z^+ \mid x,C)
}
\right)\right].
\end{equation}
Here, $x$ is the input prompt and $C$ is the structured constraint set;; $(z^+, z^-) \sim D$ are preference pairs 
with $z^+$ ranked higher than $z^-$ by deterministic reward $r(\cdot)$; 
$\pi_\theta, \pi_{\text{ref}}$ are the policy and reference models; 
$\beta$ is the temperature controlling deviation from $\pi_{\text{ref}}$; 
$\sigma$ is the sigmoid function, applied to compute preference probabilities in the objective.

\begin{figure*}[t]
\centering
\includegraphics[width=1.0\linewidth]{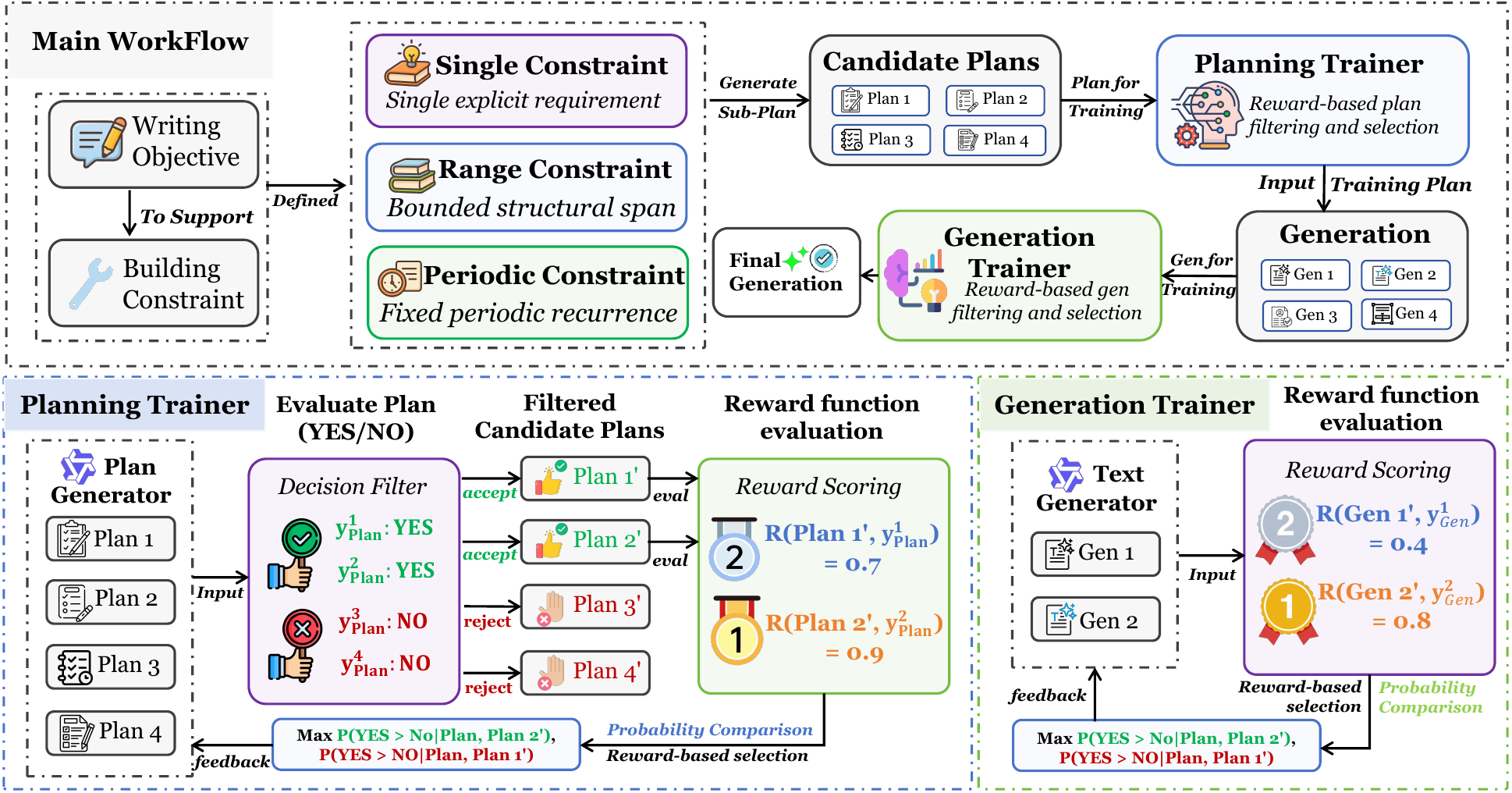}
\caption{Overview of LoopWeaver. Complete-plan construction, in-context revision, and whole-plan filtering operate at inference time; shared-reward plan-level and generation-level DPO provide sequential coupling during offline training.}
\label{F3}
\end{figure*}

\section{Methodology: LoopWeaver}
\label{sec:methodology}
As illustrated in Figure~\ref{F3}, we present \textbf{LoopWeaver}, a hierarchical workflow with three components: \emph{constraint-aware hierarchical planning}, \emph{feasibility filtering}, and \emph{reward-based preference optimization}. The first two operate in the feed-forward inference workflow, while the shared reward couples the plan-level and generation-level DPO stages during offline training.

\subsection{Constraint-Aware Hierarchical Planning}
\label{sec:plan-modeling}

LoopWeaver adopts a two-level architecture separating a \emph{plan level} that captures global structure from a \emph{generation level} that realizes each unit into text. Given an input prompt $x$ and a constraint set $C = C_{\text{once}} \cup C_{\text{range}} \cup C_{\text{periodic}}$ partitioned by pattern following LongGenBench~\citep{LongGenBench}, the framework produces an ordered sub-plan sequence and an aligned segment sequence, written as a two-level form:
\begin{equation}
\underbrace{p = (S_1,\, S_2,\, \ldots,\, S_m)}_{\text{plan level}}
\;\longrightarrow\;
\underbrace{G = (G_1,\, G_2,\, \ldots,\, G_m)}_{\text{generation level}},
\end{equation}
where each sub-plan $S_j$ expands into the corresponding segment $G_j$, and the integer $m$ denotes the number of units determined by the planner. The complete plan is generated from the prompt and structured constraint set, and the two levels are coupled by the shared reward used in the sequential offline DPO stages described in Section~\ref{sec:reward-opt}:
\begin{equation}
p \;\sim\; \pi_\theta\!\big(\,\cdot \,\big|\, x,\, C\big).
\end{equation}
The planner then checks and revises the sub-plans within the joint context of the complete draft and the original requirements. The revision returns a complete plan, so an upstream edit and any affected downstream units can be checked and adjusted together:
\begin{equation}
\label{eq:local-repair}
p' \;=\; \mathcal{R}\!\big(p;\, x, C\big).
\end{equation}
The repair operator targets individual sub-plans but receives and returns the complete plan. The revised plan $p'$ is subsequently passed to the whole-plan feasibility filter, which checks completeness, logical coherence, and contextual consistency before segment generation.
This provides plan-level coordination before generation, but it is not a formal guarantee of final document-level coherence: at inference, segments are generated sequentially in separate calls conditioned on the current sub-plan and the running narrative, and the generated segments are not returned to the planner for plan revision.

\textbf{Empirical Observation 1.} \textit{Constraint-aware hierarchical planning is associated with improved plan validity and constraint coverage under the evaluation protocol in Section~\ref{sec:experiments}.}

\subsection{Feasibility Filtering}
\label{sec:binary-filter}

Revised plans still vary in plan-level quality, so a lightweight feasibility filter discards candidates that fail task-inherent criteria before they enter the generation stage. From the candidate pool $P_{\text{final}} = \{p_1, p_2, \ldots, p_N\}$ produced by Section~\ref{sec:plan-modeling}, the same backbone receives each complete plan together with the prompt and constraint set and returns a prompted YES/NO feasibility judgment:
\begin{equation}
\delta_i \;=\; \mathbf{1}\!\left[p_{\text{filter}}(p_i, x, C)=\mathrm{YES}\right],
\qquad i = 1,\, 2,\ldots,N,
\end{equation}
where the prompt asks the model to assess six criteria: \emph{relevance, completeness, coherence, efficiency, specificity}, and \emph{consistency}. The filter produces one binary decision; it does not use a separately trained filter or multiply six probabilities. The accepted plans are gathered into the filtered set:
\begin{equation}
\label{eq:filter-set}
P_{\text{filtered}}
\;=\; \big\{\, p_i \in P_{\text{final}} \;\big|\; \delta_i = 1 \,\big\},
\end{equation}
which is forwarded to the generation-level optimization in Section~\ref{sec:reward-opt}, while rejected candidates are either discarded or regenerated under the same prompt $x$ and constraint set $C$, ensuring that deterministic reward scoring is applied only to plans that pass a coarse feasibility check.

\textbf{Empirical Observation 2.} \textit{Feasibility filtering reduces the number of malformed or misaligned plans passed to generation under the evaluation protocol in Section~\ref{sec:experiments}.}

\definecolor{myblue}{RGB}{230, 240, 255}
\definecolor{mypurple}{RGB}{240, 230, 255}

\subsection{Reward-Based Preference Optimization}
\label{sec:reward-opt}

LoopWeaver couples the two levels through a single shared reward $R(\cdot, e)$ defined in Section~\ref{sec:preliminaries}, evaluated under the context $e=(x,C)$ and used to score both sub-plans and segments. At the \emph{plan level}, each candidate sub-plan $S_j$ from a filtered plan in $P_{\text{filtered}}$ is scored under this reward:
\begin{equation}
\label{eq:plan-reward}
r(S_j)
\;=\; R\!\big(S_j,\; e\big),
\end{equation}
For two candidate sub-plans $S_j^+$ and $S_j^-$ drawn from the same context, the higher-reward candidate is preferred when $r(S_j^+) > r(S_j^-)$; the resulting preference pairs are used to train the plan policy through the plan-level DPO objective:
\begin{equation}
\mathcal{L}^{\text{plan}}_{\text{DPO}}(\theta)
\;=\; -\, \mathbb{E}_{(S_j^+,\, S_j^-) \sim D_{\text{plan}}}\!
\left[\log \sigma\!\left(
\beta \log
\frac{
\pi_\theta(S_j^+ \mid x, C)\,\pi_{\text{ref}}(S_j^- \mid x, C)
}{
\pi_\theta(S_j^- \mid x, C)\,\pi_{\text{ref}}(S_j^+ \mid x, C)
}
\right)\right],
\end{equation}
in which $D_{\text{plan}}$ collects reward-ranked plan pairs, $\pi_{\text{ref}}$ is the frozen reference policy, and $\beta$ is the DPO temperature controlling deviation from $\pi_{\text{ref}}$. At the \emph{generation level}, each accepted sub-plan $S_k$ is expanded into a segment $G_k$ and scored by the same shared reward,
$r(G_k)=R\big(G_k,e\big)$. Pairs $(G_k^+, G_k^-)$ with $r(G_k^+) > r(G_k^-)$ supervise the generation policy through the generation-level DPO objective:
\begin{equation}
\label{eq:gen-dpo}
\mathcal{L}^{\text{gen}}_{\text{DPO}}(\theta)
\;=\; -\, \mathbb{E}_{(G_k^+,\, G_k^-,\, S_k) \sim D_{\text{gen}}}\!
\left[\log \sigma\!\left(
\beta \log
\frac{
\pi_\theta(G_k^+ \mid x, C, S_k)\,\pi_{\text{ref}}(G_k^- \mid x, C, S_k)
}{
\pi_\theta(G_k^- \mid x, C, S_k)\,\pi_{\text{ref}}(G_k^+ \mid x, C, S_k)
}
\right)\right],
\end{equation}
where $D_{\text{gen}}$ collects segment-level preference triples conditioned on $S_k$. The same $R(\cdot,e)$ is used at both levels, while the two DPO stages are optimized sequentially with checkpoint transfer from the plan policy to the generation policy. This is the sense in which the stages are coupled during offline training; inference remains feed-forward.

\textbf{Empirical Observation 3.} \textit{Reward-based preference optimization improves alignment with the verifiable constraint objectives under the evaluation protocol in Section~\ref{sec:experiments}.}
% \textbf{Constraint Enforcement in LoopWeaver.} $C$ enters the framework via four mechanisms: (i) complete-plan revision (Eq.~\ref{eq:local-repair}); (ii) prompt-only feasibility screening over six task-inherent criteria (Eq.~\ref{eq:filter-set}); (iii) the $r_{\text{constraint}}$ term scoring all three constraint types in the DPO reward (Eqs.~\ref{eq:constraint-r}, \ref{eq:plan-reward}--\ref{eq:gen-dpo}); and (iv) structured prompt encoding that embeds $C$ as trackable state variables (Appendices~\ref{AppendixA1}--\ref{AppendixA2}).

% \subsection{Main Workflow}
% Figure~\ref{F3} summarizes LoopWeaver’s complete end-to-end workflow framework. During training, global plans are generated, filtered, reward-optimized, and expanded into text segments. During inference, the system generates and filters plans, executing the selected plan to produce the final output while ensuring structural coherence and constraint satisfaction.

% % Required packages
% \usepackage{booktabs}
% \usepackage{xcolor}
% \usepackage{colortbl}
% \usepackage{graphicx}

% 修复 NeurIPS 模板双括号问题，并统一字号
\captionsetup[figure]{font=small}
\captionsetup[subfigure]{
    font=small,
    labelformat=simple,
    labelsep=space,          % 括号后跟空格（原为 period，即句点）
}
\renewcommand\thesubfigure{(\alph{subfigure})}

% ==================== TABLE 1 ====================
\begin{table*}[t]
\centering
\small
\setlength{\tabcolsep}{3pt}
\renewcommand{\arraystretch}{1.18}

\caption{\label{T1}Comparison of narrative-level text quality and constraint-following accuracy across different methods.}

\resizebox{\textwidth}{!}{%
\begin{tabular}{@{}lccccccccc@{}}
\toprule
\textbf{Method}
& \multicolumn{5}{c}{\textbf{Text Quality}}
& \multicolumn{4}{c}{\textbf{Constraint Accuracy}} \\
\cmidrule(lr){2-6} \cmidrule(lr){7-10}
& \textbf{Narr.} & \textbf{Memory} & \textbf{Temporal} & \textbf{Affect.} & \textbf{Avg.}
& \textbf{Once} & \textbf{Range} & \textbf{Peri.} & \textbf{Avg.} \\
\midrule

\textbf{Qwen2.5-0.5B-Instruct}
& \underline{26.05{\scriptsize$\pm$0.26}} & 24.31{\scriptsize$\pm$0.23} & 19.74{\scriptsize$\pm$0.57} & \textbf{39.86{\scriptsize$\pm$0.39}} & \underline{27.49{\scriptsize$\pm$0.21}}
& 23.46{\scriptsize$\pm$1.59} & 16.60{\scriptsize$\pm$1.52} & 15.69{\scriptsize$\pm$0.54} & 18.58{\scriptsize$\pm$0.75} \\
\quad + CogWriter
& 23.38{\scriptsize$\pm$0.14} & 24.24{\scriptsize$\pm$0.43} & 19.51{\scriptsize$\pm$0.32} & 33.50{\scriptsize$\pm$0.26} & 25.16{\scriptsize$\pm$0.12}
& \underline{26.61{\scriptsize$\pm$0.98}} & {18.03{\scriptsize$\pm$0.30}} & \textbf{17.70{\scriptsize$\pm$0.24}} & \underline{20.78{\scriptsize$\pm$0.21}} \\
\quad + LongWriter
& 24.87{\scriptsize$\pm$0.30} & \underline{25.26{\scriptsize$\pm$0.30}} & \underline{22.41{\scriptsize$\pm$0.40}} & 31.26{\scriptsize$\pm$0.39} & 25.95{\scriptsize$\pm$0.13}
& 21.01{\scriptsize$\pm$0.30} & \underline{19.29{\scriptsize$\pm$0.11}} & \underline{15.83{\scriptsize$\pm$0.20}} & 18.71{\scriptsize$\pm$0.13} \\
\rowcolor{myblue}
\quad + LoopWeaver (ours)
& \textbf{29.17{\scriptsize$\pm$0.22}} & \textbf{35.02{\scriptsize$\pm$0.57}} & \textbf{33.80{\scriptsize$\pm$0.46}} & \underline{37.39{\scriptsize$\pm$0.20}} & \textbf{33.84{\scriptsize$\pm$0.31}}
& \textbf{28.01{\scriptsize$\pm$0.55}} & \textbf{22.52{\scriptsize$\pm$0.10}} & 15.80{\scriptsize$\pm$0.28} & \textbf{22.11{\scriptsize$\pm$0.31}} \\
\midrule

\textbf{Qwen2.5-1.5B-Instruct}
& 41.96{\scriptsize$\pm$0.42} & 40.12{\scriptsize$\pm$0.37} & 34.63{\scriptsize$\pm$0.47} & \textbf{59.17{\scriptsize$\pm$0.73}} & 43.97{\scriptsize$\pm$0.45}
& 23.30{\scriptsize$\pm$1.43} & 22.22{\scriptsize$\pm$0.28} & 12.12{\scriptsize$\pm$0.60} & 19.22{\scriptsize$\pm$0.77} \\
\quad + CogWriter
& \underline{42.32{\scriptsize$\pm$0.31}} & \underline{45.36{\scriptsize$\pm$0.64}} & 38.21{\scriptsize$\pm$0.13} & 56.00{\scriptsize$\pm$0.37} & \underline{45.47{\scriptsize$\pm$0.32}}
& 31.66{\scriptsize$\pm$3.09} & 25.76{\scriptsize$\pm$0.91} & 17.79{\scriptsize$\pm$0.22} & 25.07{\scriptsize$\pm$1.41} \\
\quad + LongWriter
& 41.32{\scriptsize$\pm$0.47} & 44.24{\scriptsize$\pm$0.28} & \underline{39.01{\scriptsize$\pm$0.53}} & 55.13{\scriptsize$\pm$0.42} & 44.93{\scriptsize$\pm$0.15}
& \underline{42.67{\scriptsize$\pm$0.36}} & \underline{28.17{\scriptsize$\pm$1.16}} & \underline{41.24{\scriptsize$\pm$0.49}} & \underline{37.36{\scriptsize$\pm$0.44}} \\
\rowcolor{myblue}
\quad + LoopWeaver (ours)
& \textbf{47.98{\scriptsize$\pm$1.00}} & \textbf{55.21{\scriptsize$\pm$0.94}} & \textbf{49.27{\scriptsize$\pm$0.83}} & \underline{56.83{\scriptsize$\pm$0.38}} & \textbf{52.32{\scriptsize$\pm$0.67}}
& \textbf{49.01{\scriptsize$\pm$0.16}} & \textbf{38.88{\scriptsize$\pm$0.13}} & \textbf{42.93{\scriptsize$\pm$0.30}} & \textbf{43.61{\scriptsize$\pm$0.20}} \\
\midrule

\textbf{Qwen2.5-7B-Instruct}
& 69.75{\scriptsize$\pm$0.66} & 70.94{\scriptsize$\pm$0.23} & 67.21{\scriptsize$\pm$0.44} & 77.46{\scriptsize$\pm$0.47} & 71.34{\scriptsize$\pm$0.34}
& 40.97{\scriptsize$\pm$0.74} & 32.86{\scriptsize$\pm$3.44} & 18.39{\scriptsize$\pm$2.22} & 30.74{\scriptsize$\pm$2.12} \\
\quad + CogWriter
& \underline{73.33{\scriptsize$\pm$0.36}} & \underline{76.02{\scriptsize$\pm$0.10}} & \underline{72.69{\scriptsize$\pm$0.22}} & \underline{80.29{\scriptsize$\pm$0.38}} & \underline{75.58{\scriptsize$\pm$0.09}}
& \underline{42.05{\scriptsize$\pm$0.91}} & \underline{56.65{\scriptsize$\pm$0.35}} & 27.90{\scriptsize$\pm$0.43} & \underline{42.20{\scriptsize$\pm$0.56}} \\
\quad + LongWriter
& 68.33{\scriptsize$\pm$0.32} & 73.68{\scriptsize$\pm$0.44} & 67.15{\scriptsize$\pm$0.90} & 79.34{\scriptsize$\pm$0.14} & 72.12{\scriptsize$\pm$0.20}
& 27.23{\scriptsize$\pm$0.17} & 32.48{\scriptsize$\pm$1.07} & \underline{27.91{\scriptsize$\pm$0.26}} & 29.21{\scriptsize$\pm$0.50} \\
\rowcolor{myblue}
\quad + LoopWeaver (ours)
& \textbf{75.98{\scriptsize$\pm$0.40}} & \textbf{79.42{\scriptsize$\pm$0.37}} & \textbf{78.73{\scriptsize$\pm$0.88}} & \textbf{80.94{\scriptsize$\pm$0.71}} & \textbf{78.77{\scriptsize$\pm$0.23}}
& \textbf{53.33{\scriptsize$\pm$0.54}} & \textbf{56.90{\scriptsize$\pm$0.51}} & \textbf{35.01{\scriptsize$\pm$1.29}} & \textbf{48.41{\scriptsize$\pm$0.75}} \\
\midrule

\textbf{LLaMA3.1-8B-Instruct}
& 39.08{\scriptsize$\pm$0.41} & 39.19{\scriptsize$\pm$0.27} & 35.56{\scriptsize$\pm$0.49} & 41.79{\scriptsize$\pm$0.07} & 38.91{\scriptsize$\pm$0.09}
& 31.51{\scriptsize$\pm$0.84} & 27.87{\scriptsize$\pm$0.43} & 21.02{\scriptsize$\pm$0.39} & 26.80{\scriptsize$\pm$0.55} \\
\quad + CogWriter
& 53.29{\scriptsize$\pm$0.32} & 57.91{\scriptsize$\pm$0.32} & 53.71{\scriptsize$\pm$0.41} & 59.67{\scriptsize$\pm$0.35} & 56.14{\scriptsize$\pm$0.12}
& \underline{43.12{\scriptsize$\pm$0.11}} & 48.77{\scriptsize$\pm$0.07} & \underline{33.80{\scriptsize$\pm$0.07}} & \underline{41.89{\scriptsize$\pm$0.08}} \\
\quad + LongWriter
& \underline{57.31{\scriptsize$\pm$0.54}} & \underline{61.62{\scriptsize$\pm$0.70}} & \underline{59.33{\scriptsize$\pm$0.69}} & \underline{60.08{\scriptsize$\pm$0.07}} & \underline{59.59{\scriptsize$\pm$0.42}}
& 41.01{\scriptsize$\pm$0.30} & \underline{49.29{\scriptsize$\pm$0.11}} & 32.83{\scriptsize$\pm$0.20} & 41.04{\scriptsize$\pm$0.20} \\
\rowcolor{myblue}
\quad + LoopWeaver (ours)
& \textbf{62.52{\scriptsize$\pm$0.50}} & \textbf{65.00{\scriptsize$\pm$0.23}} & \textbf{63.56{\scriptsize$\pm$0.71}} & \textbf{68.50{\scriptsize$\pm$0.39}} & \textbf{64.90{\scriptsize$\pm$0.42}}
& \textbf{45.94{\scriptsize$\pm$1.26}} & \textbf{53.27{\scriptsize$\pm$1.05}} & \textbf{46.44{\scriptsize$\pm$1.28}} & \textbf{48.55{\scriptsize$\pm$1.20}} \\
\midrule

\textbf{gpt-4o-mini}
& 71.81{\scriptsize$\pm$0.23} & 73.67{\scriptsize$\pm$0.53} & 64.77{\scriptsize$\pm$0.30} & 82.44{\scriptsize$\pm$0.39} & 73.17{\scriptsize$\pm$0.31}
& 51.47{\scriptsize$\pm$0.21} & 40.00{\scriptsize$\pm$0.11} & 44.36{\scriptsize$\pm$0.30} & 45.28{\scriptsize$\pm$0.13} \\
\quad + CogWriter
& {78.38{\scriptsize$\pm$0.50}} & {80.54{\scriptsize$\pm$0.50}} & {74.99{\scriptsize$\pm$1.66}} & {85.03{\scriptsize$\pm$0.55}} & {79.73{\scriptsize$\pm$0.77}}
& {65.63{\scriptsize$\pm$0.24}} & {42.16{\scriptsize$\pm$0.98}} & {32.66{\scriptsize$\pm$0.20}} & {46.82{\scriptsize$\pm$0.30}} \\
\bottomrule
\end{tabular}
}
\end{table*}

% ==================== TABLE 2 ====================
\begin{table*}[t]
\centering
\small
\setlength{\tabcolsep}{3pt}
\renewcommand{\arraystretch}{1.18}

\caption{\label{T2}Ablation study of Planning and Generation modules in LoopWeaver (Qwen2.5-7B).}

\resizebox{\textwidth}{!}{%
\begin{tabular}{@{}lccccccccc@{}}
\toprule
\textbf{Method}
& \multicolumn{5}{c}{\textbf{Text Quality}}
& \multicolumn{4}{c}{\textbf{Constraint Accuracy}} \\
\cmidrule(lr){2-6} \cmidrule(lr){7-10}
& \textbf{Narr.} & \textbf{Memory} & \textbf{Temporal}
& \textbf{Affect.} & \textbf{Avg.}
& \textbf{Once} & \textbf{Range} & \textbf{Peri.} & \textbf{Avg.} \\
\midrule

Base Model
& 69.75{\scriptsize$\pm$0.66}
& 70.94{\scriptsize$\pm$0.23}
& 67.21{\scriptsize$\pm$0.44}
& 77.46{\scriptsize$\pm$0.47}
& 71.34{\scriptsize$\pm$0.35}
& 40.97{\scriptsize$\pm$0.74}
& 32.86{\scriptsize$\pm$3.44}
& 18.39{\scriptsize$\pm$2.22}
& 30.74{\scriptsize$\pm$2.12} \\

+ Planning
& \underline{74.73{\scriptsize$\pm$0.16}}
& \underline{77.81{\scriptsize$\pm$0.47}}
& \underline{76.60{\scriptsize$\pm$0.58}}
& \underline{79.77{\scriptsize$\pm$0.56}}
& \underline{77.23{\scriptsize$\pm$0.21}}
& 48.55{\scriptsize$\pm$0.08}
& \underline{53.63{\scriptsize$\pm$0.38}}
& 30.44{\scriptsize$\pm$0.21}
& 44.20{\scriptsize$\pm$0.05} \\

+ Generation
& 73.67{\scriptsize$\pm$0.24}
& 77.19{\scriptsize$\pm$0.60}
& 75.85{\scriptsize$\pm$0.30}
& 79.56{\scriptsize$\pm$0.43}
& 76.57{\scriptsize$\pm$0.31}
& \underline{49.59{\scriptsize$\pm$0.05}}
& 53.19{\scriptsize$\pm$0.19}
& \underline{31.55{\scriptsize$\pm$0.35}}
& \underline{44.78{\scriptsize$\pm$0.12}} \\

\rowcolor{mypurple}
Full LoopWeaver
& \textbf{75.98{\scriptsize$\pm$0.40}}
& \textbf{79.42{\scriptsize$\pm$0.37}}
& \textbf{78.73{\scriptsize$\pm$0.88}}
& \textbf{80.94{\scriptsize$\pm$0.71}}
& \textbf{78.77{\scriptsize$\pm$0.23}}
& \textbf{53.33{\scriptsize$\pm$0.54}}
& \textbf{56.90{\scriptsize$\pm$0.51}}
& \textbf{35.01{\scriptsize$\pm$1.29}}
& \textbf{48.41{\scriptsize$\pm$0.75}} \\

\bottomrule
\end{tabular}
}
\end{table*}

\section{Experiments}
\label{sec:experiments}
This section presents the experimental setup, main results, and in-depth analysis of our approach. To systematically evaluate LoopWeaver, we address the following research questions (RQs): \textbf{RQ1:} How much does LoopWeaver improve overall model performance over existing baseline approaches? \textbf{RQ2:} How effective are LoopWeaver's key components, as shown by comprehensive ablation studies? \textbf{RQ3:} How does constraint-aware workflow design enhance constraint satisfaction? \textbf{RQ4:} How does the offline preference-optimization workflow trade off inference efficiency against constraint-following accuracy? \textbf{RQ5:} How does offline plan-level and generation-level preference optimization coordinate the two levels? \textbf{RQ6:} How does backbone model scale affect constrained long-form generation under identical prompts and constraints?
\subsection{Experimental Setup}
 
\textbf{Baseline.} We compare LoopWeaver with CogWriter~\citep{WanMH0GC25} and LongWriter~\citep{bai2024longwriter} for the Qwen2.5~\citep{Qwen2.5} and LLaMA3.1-8B backbones, and we additionally include GPT-4o-mini and its CogWriter variant
as proprietary-model baselines. Details are provided in Appendix~\ref{Baseline}.

\textbf{Evaluation Metrics.} 
We evaluate model performance on LongGenBench along two axes of evaluation. Instruction Following Accuracy measures compliance with single, range-based, and periodic constraints, reporting both category-wise and overall accuracy. Text Quality is assessed across narrative coherence, memory consistency, temporal grounding, and affective consistency, using an aggregated score. Additional details are provided in Appendix~\ref{Evaluation} for completeness and reproducibility.

\textbf{Implementation Details.} We implement LoopWeaver following the \textbf{LongGenBench}~\citep{LongGenBench} framework, incorporating components for training and optimization. All experiments run on 3 NVIDIA A40 GPUs (48GB), enabling scalable distributed computation. Details are in Appendix~\ref{Implementation}.

\subsection{Reward and Feasibility Audits}
\label{sec:audits}

We audited the fixed candidate set used for preference construction rather than relying on a sampled reward estimate. The audit covers 984 records and 63,972 candidate units; recomputing the deterministic reward with the default setting $(\tau,\alpha)=(0.7,0.5)$ matches the saved value for all 984 records. For each sensitivity setting, selection retention compares the candidate identities selected for preference construction with those selected under the default setting. A record is eligible when its reward gap is at least 0.2; the eligible $n$ column reports the number and percentage of eligible records, and $\Delta n$ is measured relative to the default setting.

\begin{table*}[h]
\centering
\caption{Reward and feasibility audits. The left panel reports the fixed-candidate sensitivity audit under the deterministic reward, while the right panel reports feasibility-filter outcomes and downstream DPO-pair yield.}
\label{tab:reward-feasibility-audits}

\begin{minipage}[t]{0.49\textwidth}
\vspace{0pt}
\centering
\scriptsize
\textbf{(a) Fixed-candidate sensitivity audit}

\vspace{0.4em}

\setlength{\tabcolsep}{2pt}
\renewcommand{\arraystretch}{1.6}
\resizebox{\linewidth}{!}{%
\begin{tabular}{@{}lrrrrr@{}}
\toprule
Setting & Value & Selection retained (\%) & Selection changed (\%) & Eligible $n$ (\%) & $\Delta n$ \\
\midrule
$\tau$ ($\alpha=0.5$) & 0.6
& 97.6 & 2.4 & 924 (93.9) & +16 \\

$\tau$ ($\alpha=0.5$) & \textbf{0.7}
& \textbf{100.0} & 0.0 & 908 (92.3) & 0 \\

$\tau$ ($\alpha=0.5$) & 0.8
& 99.2 & 0.8 & 902 (91.7) & $-$6 \\

$\alpha$ ($\tau=0.7$) & 0.3
& 99.8 & 0.2 & 888 (90.2) & $-$20 \\

$\alpha$ ($\tau=0.7$) & \textbf{0.5}
& \textbf{100.0} & 0.0 & 908 (92.3) & 0 \\

$\alpha$ ($\tau=0.7$) & 0.7
& 99.0 & 1.0 & 920 (93.5) & +12 \\
\bottomrule
\end{tabular}
}
\end{minipage}\hfill%
\begin{minipage}[t]{0.49\textwidth}
\vspace{0pt}
\centering
\scriptsize
\textbf{(b) Feasibility-filter audit}

\vspace{0.4em}

\setlength{\tabcolsep}{2pt}
\resizebox{\linewidth}{!}{%
\begin{tabular}{@{}lrrr@{}}
\toprule
Audit item & Split A & Split B & Overall \\
\midrule
Candidates & 400 & 400 & 800 \\

Accepted & 251 & 241 & 492 (61.50\%) \\

Rejected & 147 & 158 & 305 (38.13\%) \\

Failed & 2 & 1 & 3 (0.38\%) \\

Valid DPO pairs from accepted plans
& 220 & 211 & 431 \\

Mean/min reward gap
& 0.93/0.50 & 0.94/0.50 & 0.94/0.50 \\

Proxy precision/recall (\%)
& 62.6/74.4 & 59.3/71.1 & 61.0/72.8 \\

Proxy false-rejection (\%)
& 25.6 & 28.9 & 27.2 \\
\bottomrule
\end{tabular}
}
\end{minipage}

\end{table*}

The feasibility-filter audit covers two 400-candidate splits. The filter uses the same Llama-3.1-8B-Instruct base checkpoint used to initialize the corresponding planning and generation policies; under the default inference wrapper, it makes one prompted YES/NO decision per complete plan with temperature 0.7, top-p 0.9, and max tokens 1200. Its structural proxy checks requested unit counts and complete Week/Floor/Block/Unit identifiers; it is not a semantic six-criterion accuracy measure. Accepted and rejected counts refer to the filter decision. The three failed cases did not produce a valid decision and are therefore excluded from the confusion-matrix metrics; over the remaining 797 cases, pooled TP/FP/FN/TN are 300/192/112/193. Proxy precision and recall are computed over these non-failed decisions, and false-rejection is defined as $1-\mathrm{recall}$. An accepted plan contributes a DPO pair only when an aligned same-context comparator with a strictly different reward is available. Thus, 492 accepted plans yield 431 valid pairs; the remaining 61 accepted plans lack an eligible aligned comparator and do not indicate additional filter failures.

% \begin{table}[h]
% \centering
% \small
% \setlength{\tabcolsep}{3pt}
% \caption{Feasibility-filter audit across two 400-candidate splits. Proxy precision, recall, and false-rejection rates are computed over non-failed filter decisions.}
% \label{tab:filter-audit}
% \begin{tabular}{@{}lrrr@{}}
% \toprule
% Audit item & Split A & Split B & Overall \\
% \midrule
% Candidates & 400 & 400 & 800 \\
% Accepted & 251 & 241 & 492 (61.50\%) \\
% Rejected & 147 & 158 & 305 (38.13\%) \\
% Failed & 2 & 1 & 3 (0.38\%) \\
% DPO pairs from accepted plans & 220 & 211 & 431 \\
% Mean/min reward gap & 0.93/0.50 & 0.94/0.50 & 0.94/0.50 \\
% Proxy precision/recall (\%) & 62.6/74.4 & 59.3/71.1 & 61.0/72.8 \\
% Proxy false-rejection (\%) & 25.6 & 28.9 & 27.2 \\
% \bottomrule
% \end{tabular}
% \end{table}

\subsection{Main Results (RQ1)}
\begin{figure*}[h]
    \centering
    
    \begin{subfigure}[b]{0.32\textwidth}
        \centering
        \includegraphics[width=\linewidth]{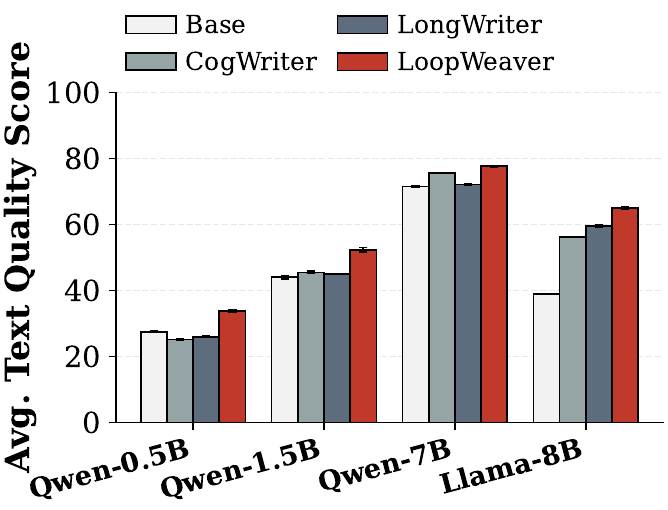}
        \vspace{-1em}
        \caption{Text Quality Evaluation}
        \label{fig:text_quality}
    \end{subfigure}
    \hfill
    \begin{subfigure}[b]{0.32\textwidth}
        \centering
        \includegraphics[width=\linewidth]{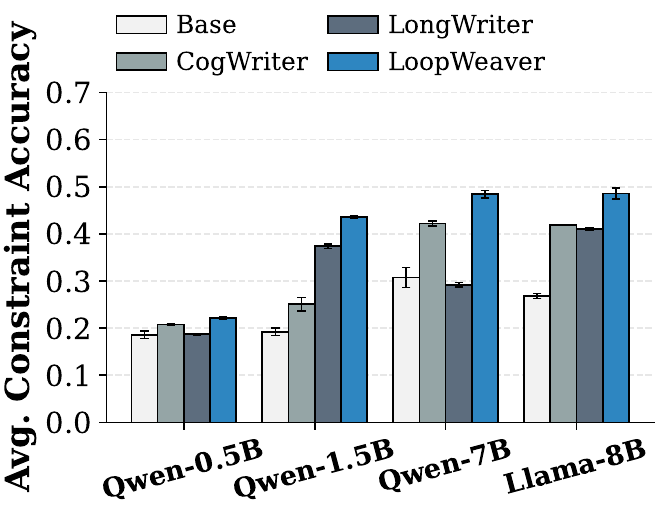}
        \vspace{-1em}
        \caption{Constraint Accuracy}
        \label{fig:constraint_accuracy}
    \end{subfigure}
    \hfill
    % \begin{subfigure}[b]{0.32\textwidth}
    %     \centering
    %     \includegraphics[width=\linewidth]{RQ3.pdf}
    %     \vspace{-1em}
    %     \caption{Ablation on Constraints}
    %     \label{fig:rq3_ablation}
    % \end{subfigure}
    \begin{subfigure}[b]{0.32\textwidth}
        \centering
        \includegraphics[width=\linewidth]{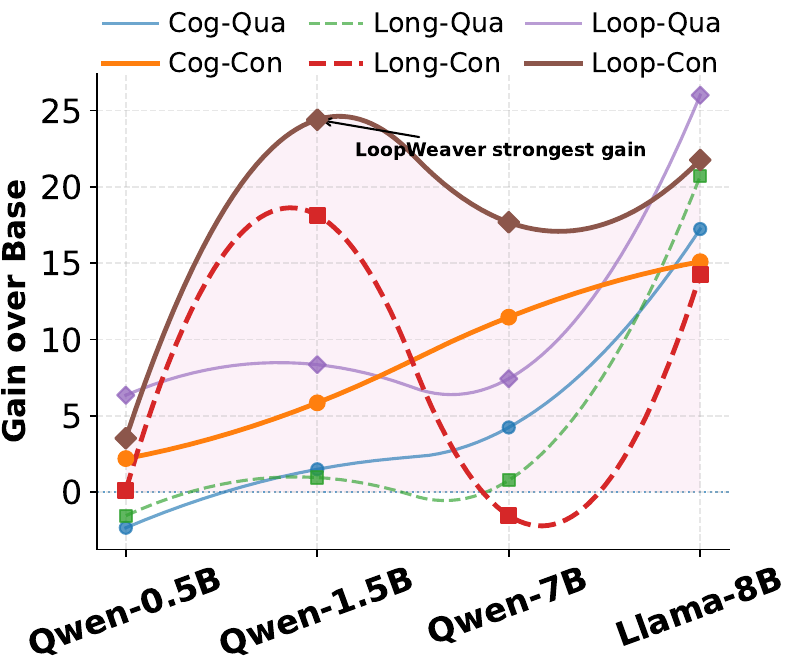}
        \vspace{-1em}
        \caption{Constraint Gain Analysis}
        \label{fig:constraint_gain}
    \end{subfigure}
    
    \caption{Comparison of LoopWeaver against baseline methods on four backbones of varying scales. \textbf{(a)} Text quality scores. \textbf{(b)} Constraint accuracy. \textbf{(c)} Gains over Base in quality and constraint accuracy across backbones.}
    
    \label{fig:F4_top}
\end{figure*}

As shown in Table~\ref{T1} and Figure~\ref{fig:F4_top}, we compare LoopWeaver with the base model, CogWriter, and LongWriter across backbones on (a) text quality and (b) constraint-following accuracy.

Across all backbones, LoopWeaver achieves superior performance on both 
metrics, jointly optimizing generation quality and constraint satisfaction. \textbf{(a) Text Quality.}
LoopWeaver achieves the highest average text-quality score among
the compared methods on all four open-source backbones, reaching
33.84, 52.32, 78.77, and 64.90 for Qwen2.5-0.5B, Qwen2.5-1.5B,
Qwen2.5-7B, and LLaMA3.1-8B, respectively. These results exceed
CogWriter by 8.68, 6.85, 3.19, and 8.76 points, demonstrating
improvements across different model scales. 
\textbf{(b) Constraint-Following Accuracy.} LoopWeaver also achieves the highest overall constraint accuracy within each backbone group. On Qwen2.5-7B, its average accuracy reaches 48.41, outperforming the base model and CogWriter by 17.67 and 6.21 percentage points, respectively. Together with the improvements on the other backbones, these results support the effectiveness of LoopWeaver for constrained long-form generation. \textbf{(c) Gain over Base.} Figure~\ref{fig:constraint_gain} shows that LoopWeaver delivers the largest gains over the corresponding base model in both average text quality and overall constraint accuracy across the four backbones. These improvements demonstrate the benefits of the proposed workflow on both evaluation axes.

\subsection{Ablation Study on the Effectiveness of LoopWeaver’s Key Components (RQ2)}

\begin{figure*}[h]
    \centering
    
    \begin{subfigure}[b]{0.24\textwidth}
        \centering
        \includegraphics[width=\linewidth]{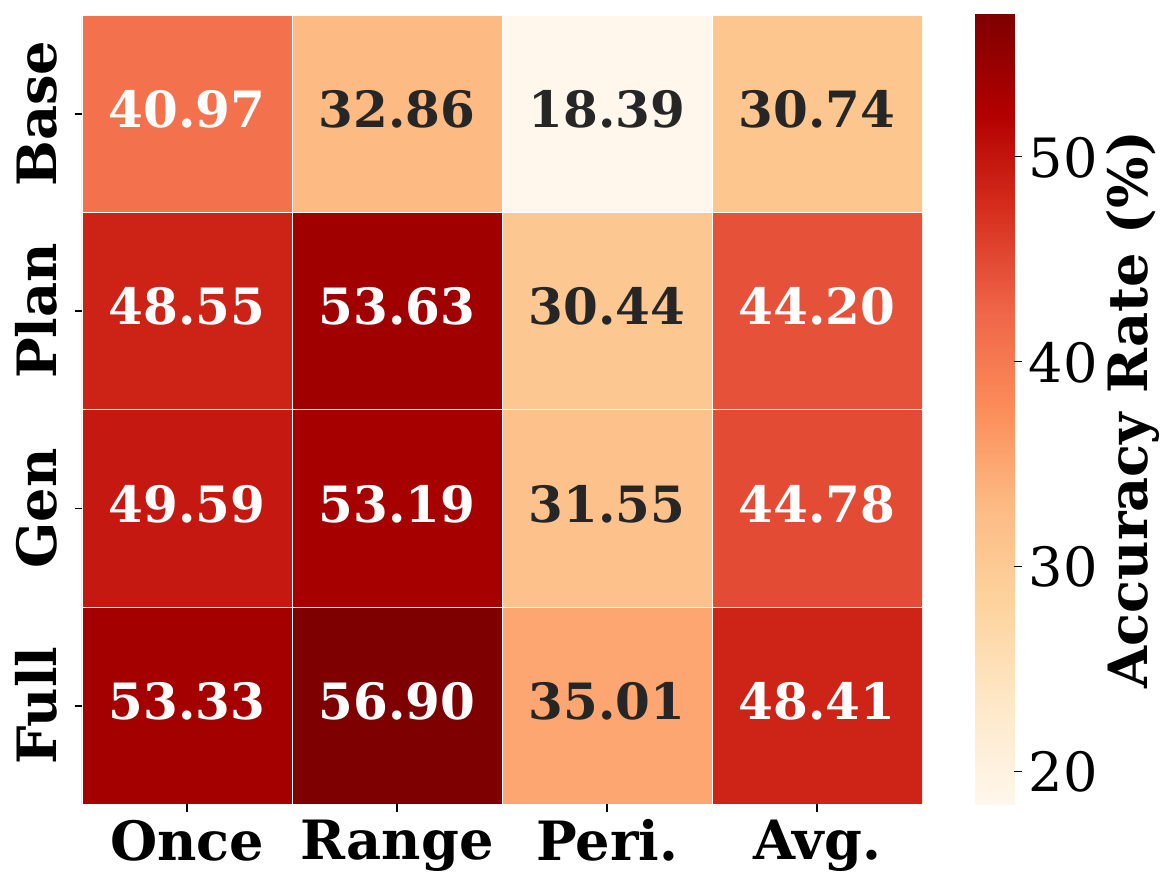}
        \vspace{-1em}
        \caption{Qwen Constraint}
        \label{fig:qwen_con}
    \end{subfigure}
    \hfill
    \begin{subfigure}[b]{0.24\textwidth}
        \centering
        \includegraphics[width=\linewidth]{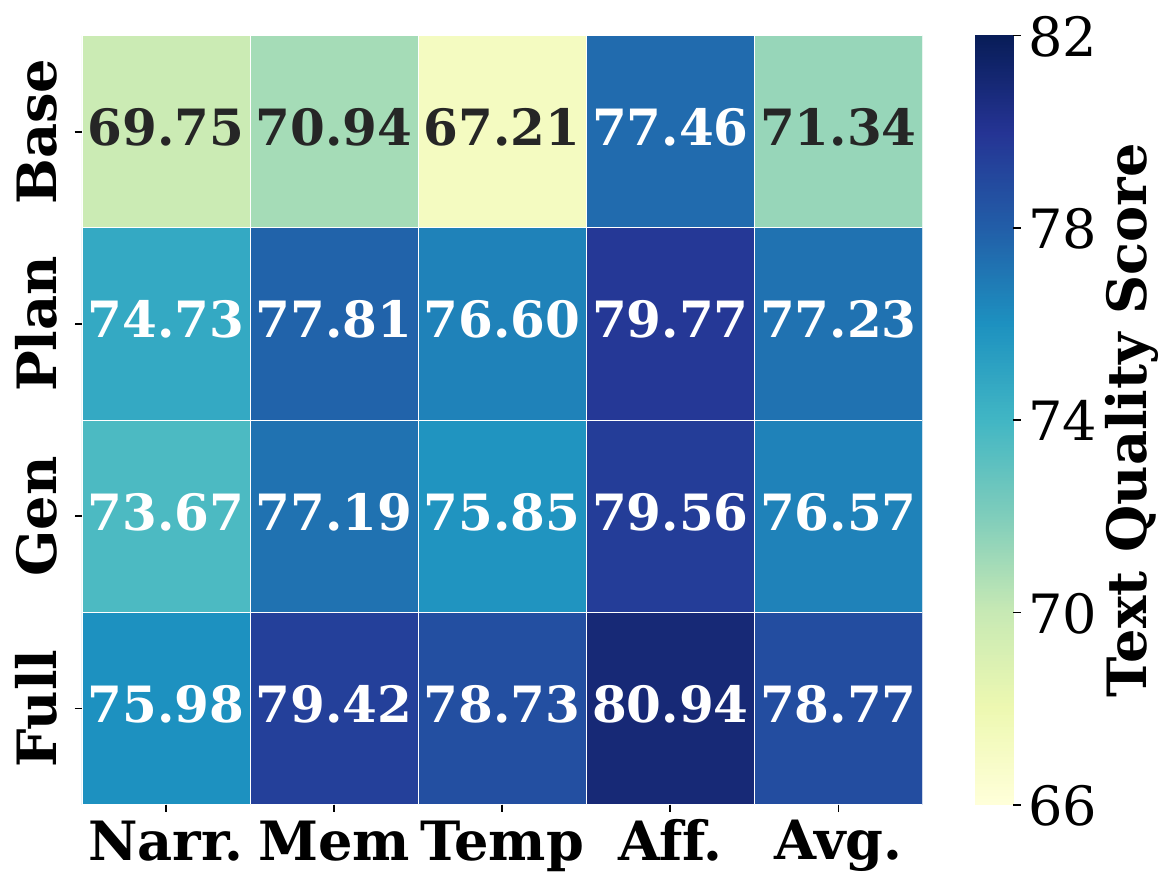}
        \vspace{-1em}
        \caption{Qwen Quality}
        \label{fig:qwen_qua}
    \end{subfigure}
    \hfill
    \begin{subfigure}[b]{0.24\textwidth}
        \centering
        \includegraphics[width=\linewidth]{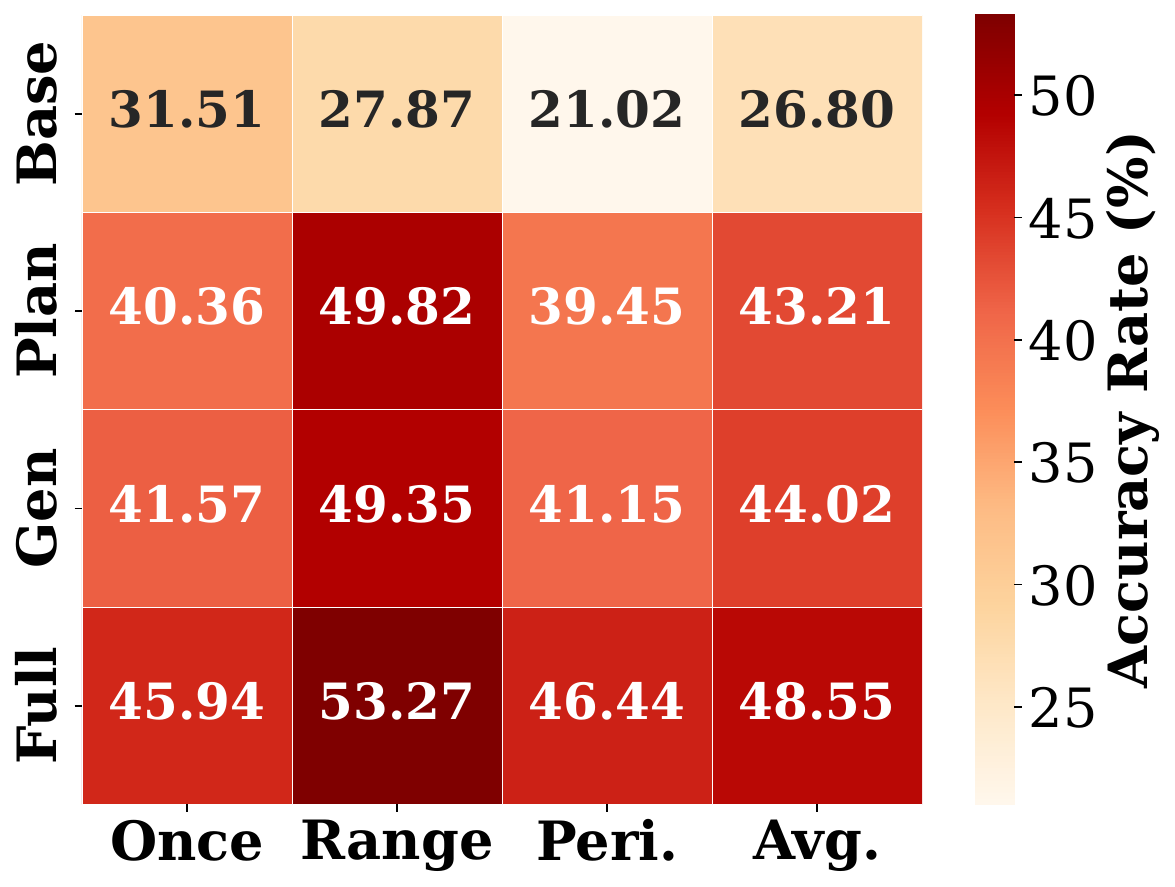}
        \vspace{-1em}
        \caption{LLaMA Constraint}
        \label{fig:llama_con}
    \end{subfigure}
    \hfill
    \begin{subfigure}[b]{0.24\textwidth}
        \centering
        \includegraphics[width=\linewidth]{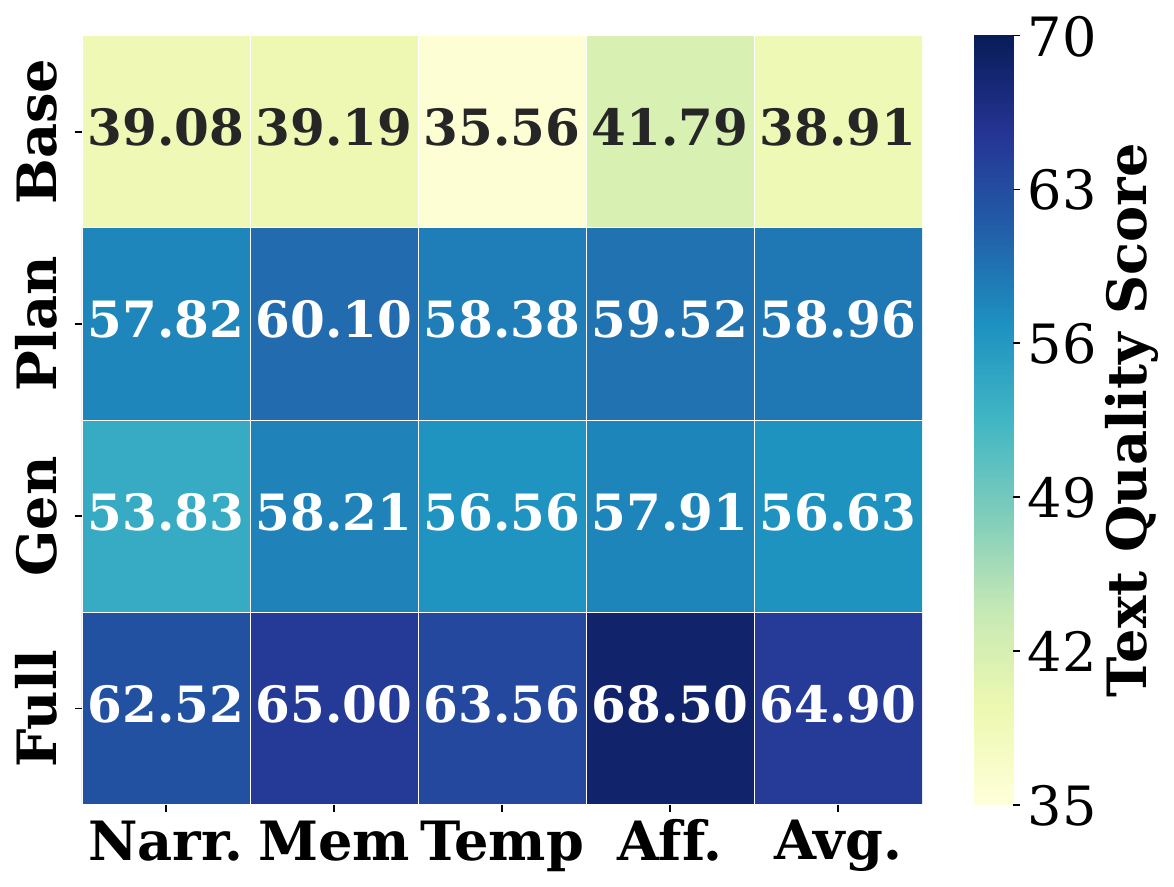}
        \vspace{-1em}
        \caption{LLaMA Quality}
        \label{fig:llama_qua}
    \end{subfigure}
    
    \caption{Ablation of LoopWeaver's Planning and Generation modules on Qwen2.5-7B and LLaMA3.1-8B backbones. \textbf{(a, c)} Constraint accuracy across constraint types. \textbf{(b, d)} Text quality across evaluation dimensions.}
    
    \label{fig:F4_bottom}
\end{figure*}

Table~\ref{T2} reports the ablation study on Qwen2.5-7B-Instruct; Figure~\ref{fig:F4_bottom} additionally reports the corresponding results for LLaMA3.1-8B.

\textbf{Both Optimization Stages Improve Constraint Satisfaction.}
As shown in Figures~\ref{fig:qwen_con} and~\ref{fig:llama_con},
training either the Planning or Generation module improves
constraint-following accuracy over the base model on both backbones.
Their relative advantages vary across constraint types, while
the full workflow achieves the highest overall accuracy,
supporting the benefits of combining planning and generation optimization.

\textbf{Sequential Plan--Generation Optimization.}
The full workflow yields the strongest constraint accuracy in this ablation, while the separate Planning-only and Generation-only variants show complementary gains across the two backbones. This supports the role of sequential plan- and generation-level optimization without implying joint backpropagation.

\textbf{Improved Text Quality through Combined Optimization.}
As shown in Figures~\ref{fig:qwen_qua} and~\ref{fig:llama_qua},
both single-stage variants improve average text quality over Base,
with Planning-only outperforming Generation-only on both backbones.
Full LoopWeaver further improves the average score to 78.77 on
Qwen2.5-7B and 64.90 on LLaMA3.1-8B, exceeding the stronger
single-stage variant by 1.54 and 5.94 points, respectively.

\subsection{Constraint-Aware Workflow Design (RQ3)}
\label{5.4}

\begin{figure*}[h]
    \centering
    \small
    
    \begin{subfigure}[b]{0.30\textwidth}
        \centering
        \includegraphics[width=\linewidth]{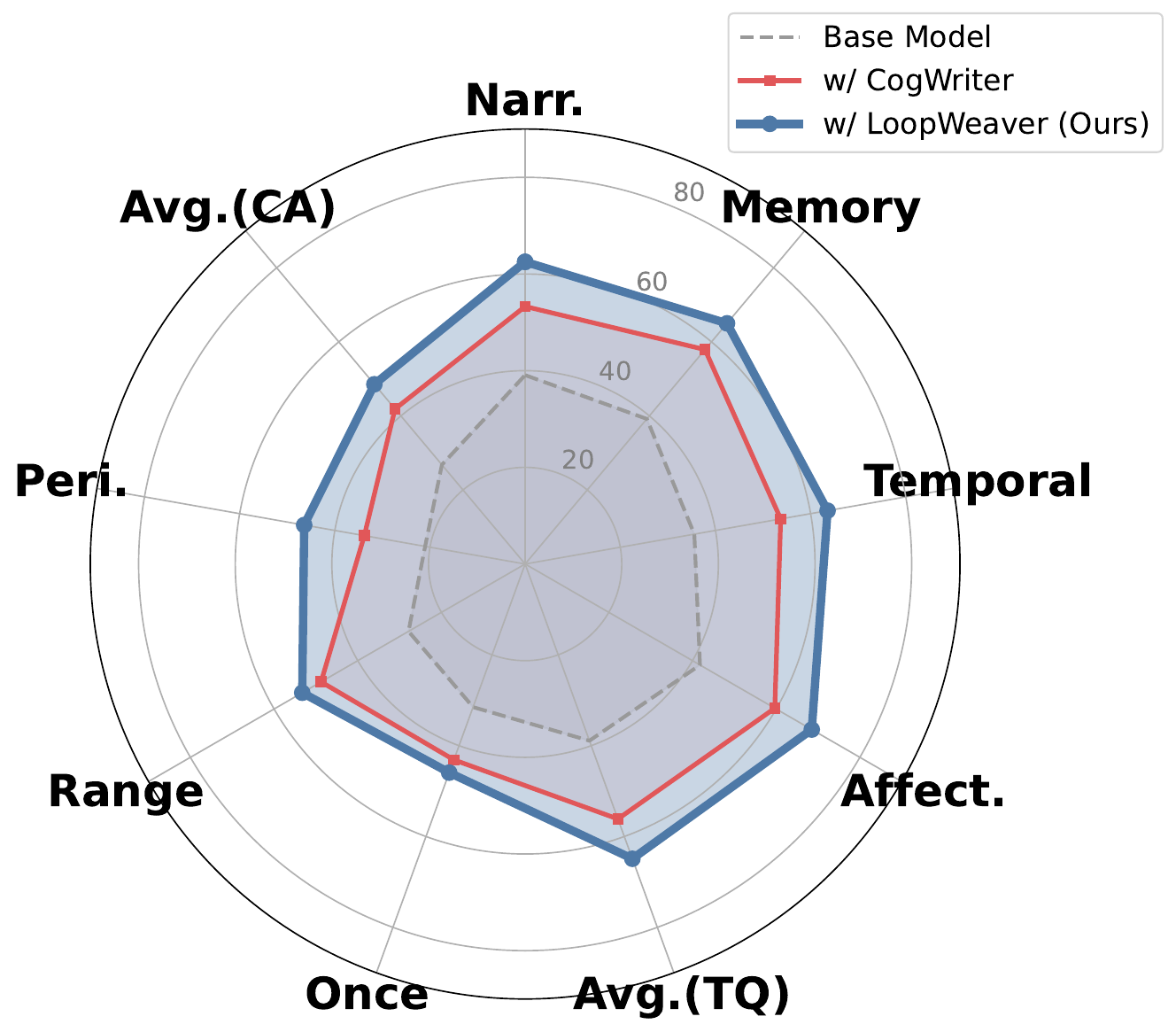}
        \vspace{-1.5em}
        \caption{Performance comparison.}
        \label{fig:rq_radar}
    \end{subfigure}
    \hfill
    \begin{subfigure}[b]{0.33\textwidth}
        \centering
        \includegraphics[width=\linewidth]{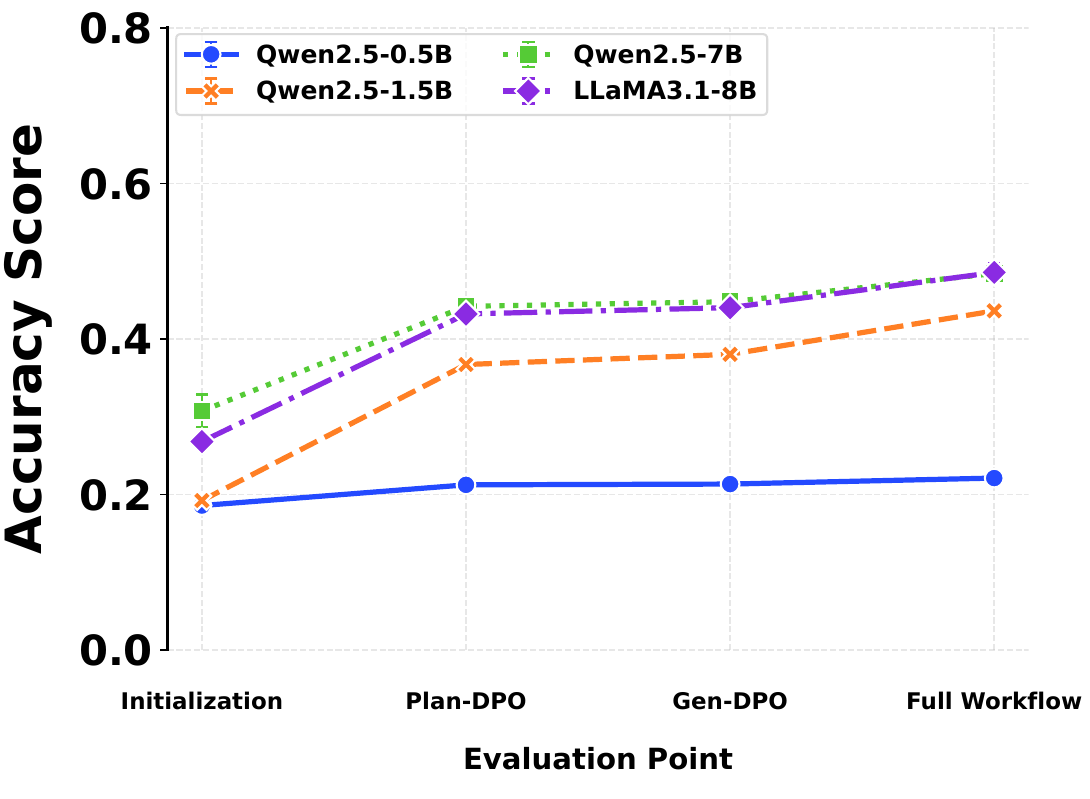}
        \vspace{-1.0em}
        \caption{Accuracy at different stages.}
        \label{fig:rq5}
    \end{subfigure}
    \hfill
    \begin{subfigure}[b]{0.33\textwidth}
        \centering
        \includegraphics[width=\linewidth]{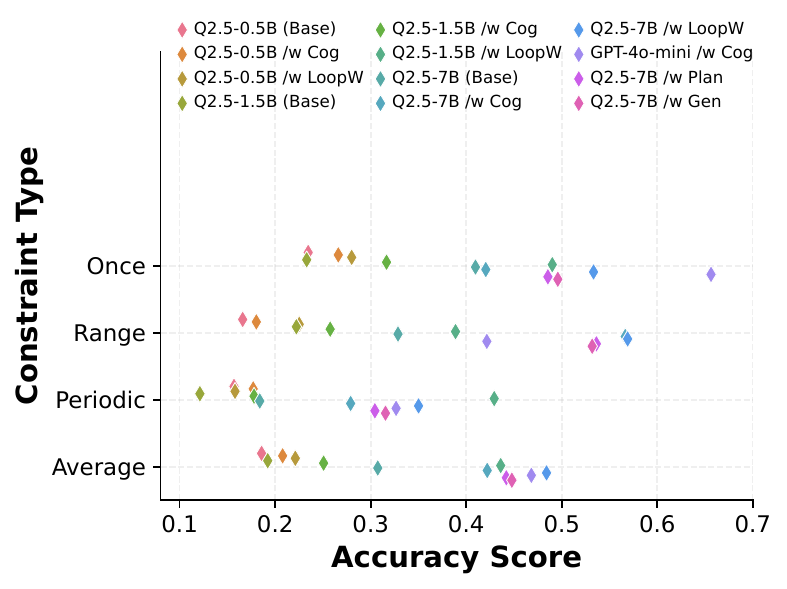}
        \vspace{-1em}
        \caption{Ablation on Constraints}
        \label{fig:rq3_ablation}
    \end{subfigure}
    
    \caption{Ablation analysis of LoopWeaver across multiple perspectives. \textbf{(a)} Multi-dimensional performance comparison. \textbf{(b)} Cumulative constraint-accuracy effects at four offline evaluation points, namely Initialization, Plan-DPO, Gen-DPO, and Final Workflow, rather than stages of one inference trajectory. \textbf{(c)} Accuracy distribution across constraint types.}
    
    \label{fig:F6_analysis}
\end{figure*}

As shown in Figure~\ref{fig:F6_analysis}, constraint-aware workflow design improves performance across multiple dimensions: overall capability, workflow-variant accuracy, and different constraint types.

\textbf{Holistic Performance Across Dimensions.}
Figure~\ref{fig:F6_analysis}(a) compares Base, CogWriter, and
LoopWeaver on LLaMA3.1-8B. LoopWeaver achieves higher scores
across the displayed text-quality dimensions and constraint-accuracy
categories than both comparison methods.

\textbf{Workflow Integration Improves Constraint Accuracy.}
Figure~\ref{fig:rq5} reports four cumulative offline evaluation points: \emph{Initialization} uses the base model; \emph{Plan-DPO} adds plan-level preference optimization; \emph{Gen-DPO} adds the sequential generation-level DPO stage; and \emph{Final Workflow} additionally enables complete-plan generation, revision, and whole-plan filtering. These points describe component additions, not an inference-time stage trajectory or iterative feedback loop.

\textbf{Constraint-Type-Specific Gains.}
Figure~\ref{fig:rq3_ablation} shows higher accuracy for LoopWeaver on several constraint types, while partial variants underperform on some settings. These results support the benchmark-specific benefit of constraint-aware design, rather than unrestricted generalization.

\subsection{Inference Efficiency with Offline Preference Optimization (RQ4)}
\label{5.5}

\begin{figure*}[h]
    \centering
    \small
    
    \begin{subfigure}[b]{0.5\textwidth}
        \centering
        \includegraphics[width=\linewidth]{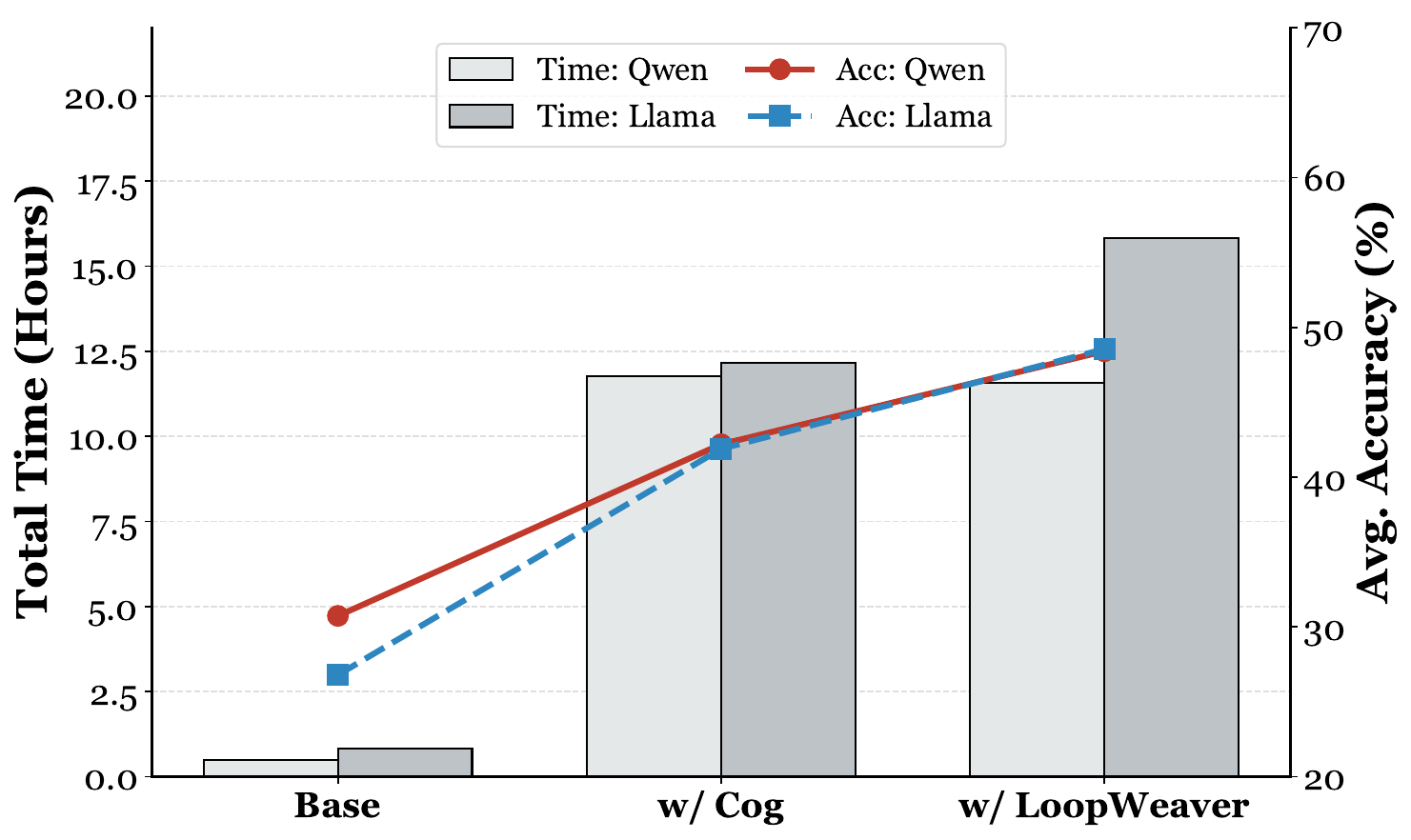}
        \vspace{-1.0em}
        \caption{Efficiency--accuracy trade-off.}
        \label{fig:rq4}
    \end{subfigure}
    \hfill
    \begin{subfigure}[b]{0.48\textwidth}
        \centering
        \includegraphics[width=\linewidth]{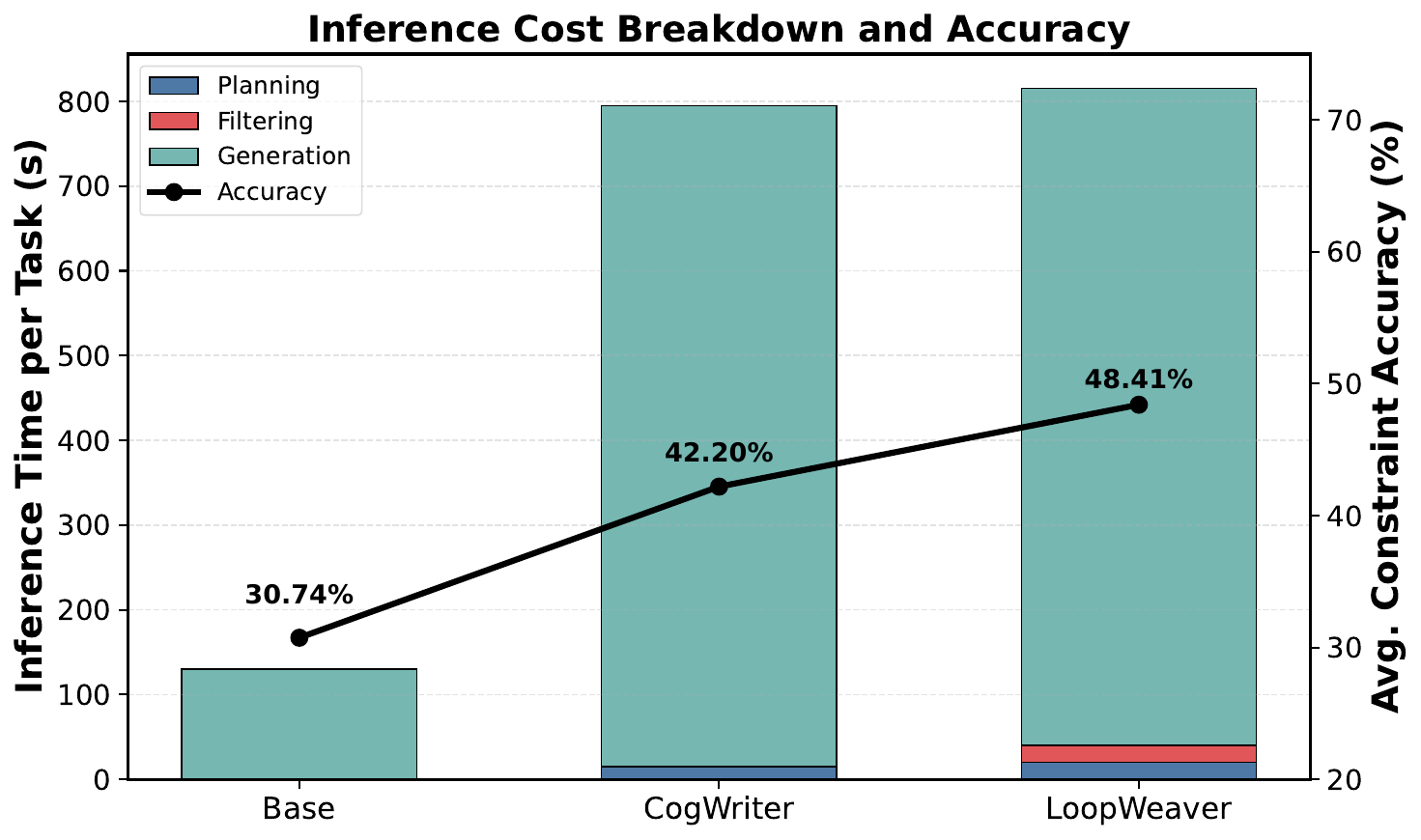}
        \vspace{-1.0em}
        \caption{Audited inference-cost breakdown and overall constraint accuracy across methods.}
        \label{fig:rq7b}
    \end{subfigure}
    % \hfill
    % \begin{subfigure}[b]{0.30\textwidth}
    %     \centering
    %     \includegraphics[width=\linewidth]{RQ7_c.pdf}
    %     \vspace{-1.5em}
    %     \caption{Performance comparison.}
    %     \label{fig:rq_radar}
    % \end{subfigure}
    
    \caption{Efficiency analysis of LoopWeaver against baselines on two backbones: \textbf{(a)} efficiency--accuracy trade-off. \textbf{(b)} audited inference-cost breakdown and overall constraint-accuracy comparison across methods; this panel is a method comparison, not an inference-stage trajectory.}    
    
    \label{fig:rq4_efficiency}
\end{figure*}

Figure~\ref{fig:rq4_efficiency} examines the efficiency-accuracy trade-off and inference cost breakdown on Qwen2.5-7B and LLaMA3.1-8B, decomposing the cost into Planning, Filtering, and Generation stages.

\textbf{Improved Accuracy with Moderate Overhead.}
LoopWeaver improves constraint-following accuracy across Qwen2.5-7B and LLaMA3.1-8B over the base model and CogWriter on the audited settings (Figure~\ref{fig:rq4}). Compared to CogWriter's static pipeline, LoopWeaver adds complete-plan construction and whole-plan filtering on the inference path, while preference-data construction and the two DPO stages are offline; the audited phase-wise costs are reported in Appendix~\ref{Overhead} and Table~\ref{tab:cost-audit}.

\subsection{Offline Preference Coordination (RQ5)}
\label{5.6}

\begin{figure*}[h]
    \centering
    \small
    
    \begin{subfigure}[b]{0.24\textwidth}
        \centering
        \includegraphics[width=\linewidth]{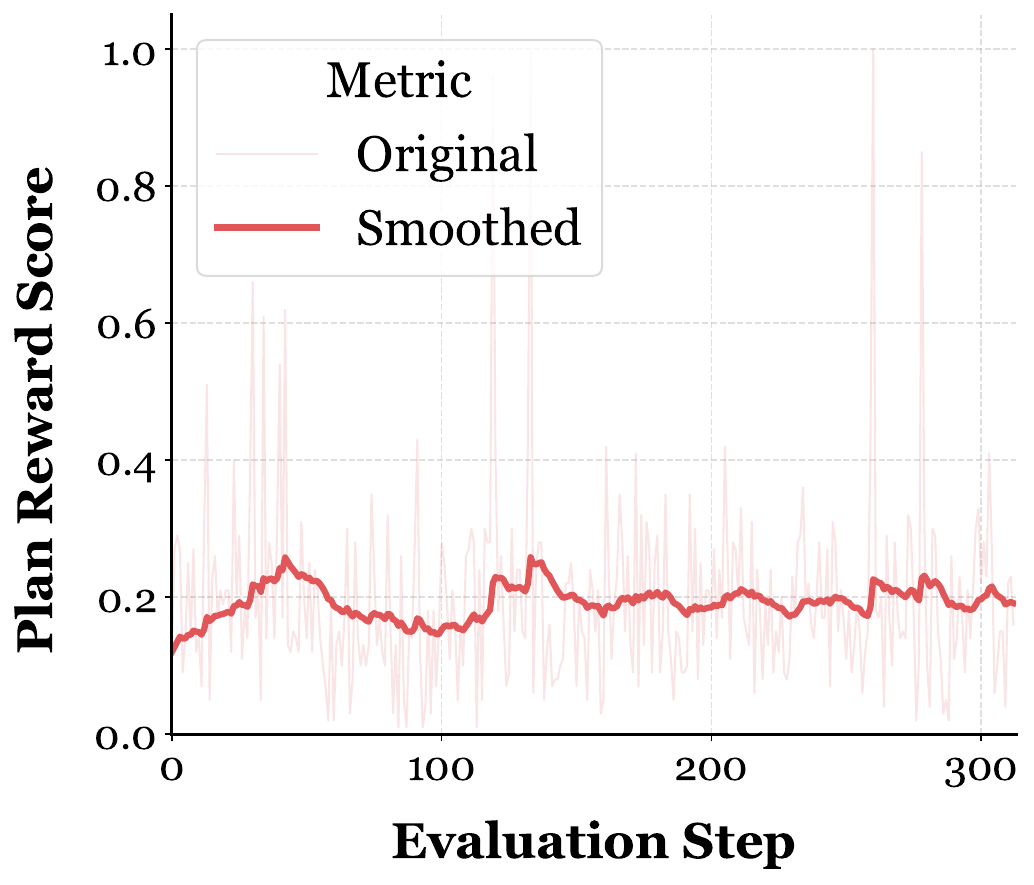}
        \vspace{-1em}
        \caption{LLaMA3.1-8B}
        \label{F6_1}
    \end{subfigure}
    \hfill
    \begin{subfigure}[b]{0.24\textwidth}
        \centering
        \includegraphics[width=\linewidth]{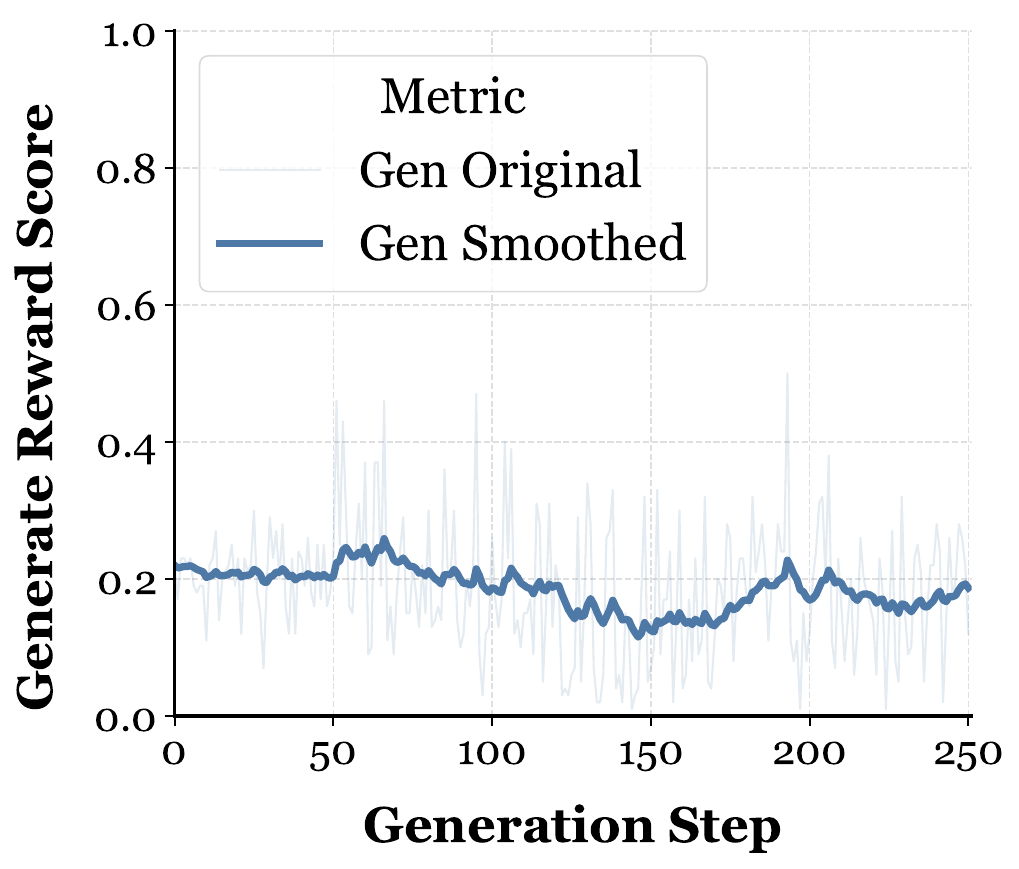}
        \vspace{-1em}
        \caption{LLaMA3.1-8B}
        \label{F6_2}
    \end{subfigure}
    \hfill
    \begin{subfigure}[b]{0.24\textwidth}
        \centering
        \includegraphics[width=\linewidth]{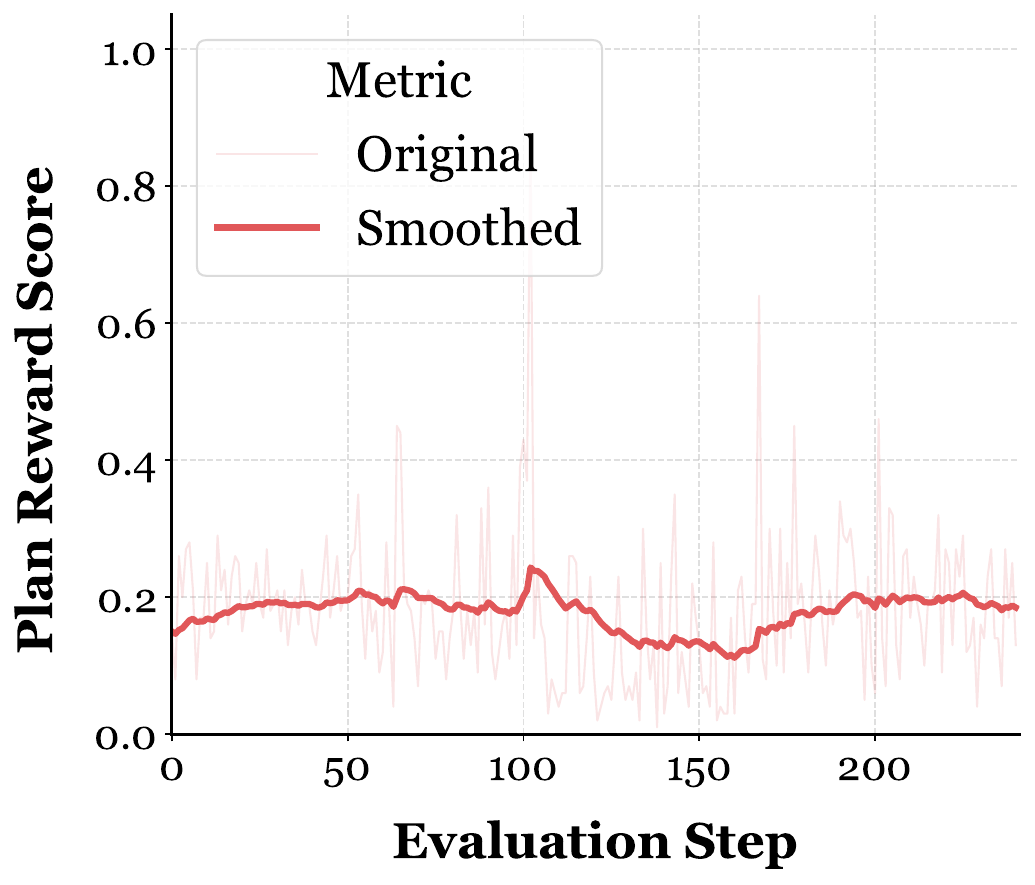}
        \vspace{-1em}
        \caption{Qwen2.5-7B}
        \label{F6_7}
    \end{subfigure}
    \hfill
    \begin{subfigure}[b]{0.24\textwidth}
        \centering
        \includegraphics[width=\linewidth]{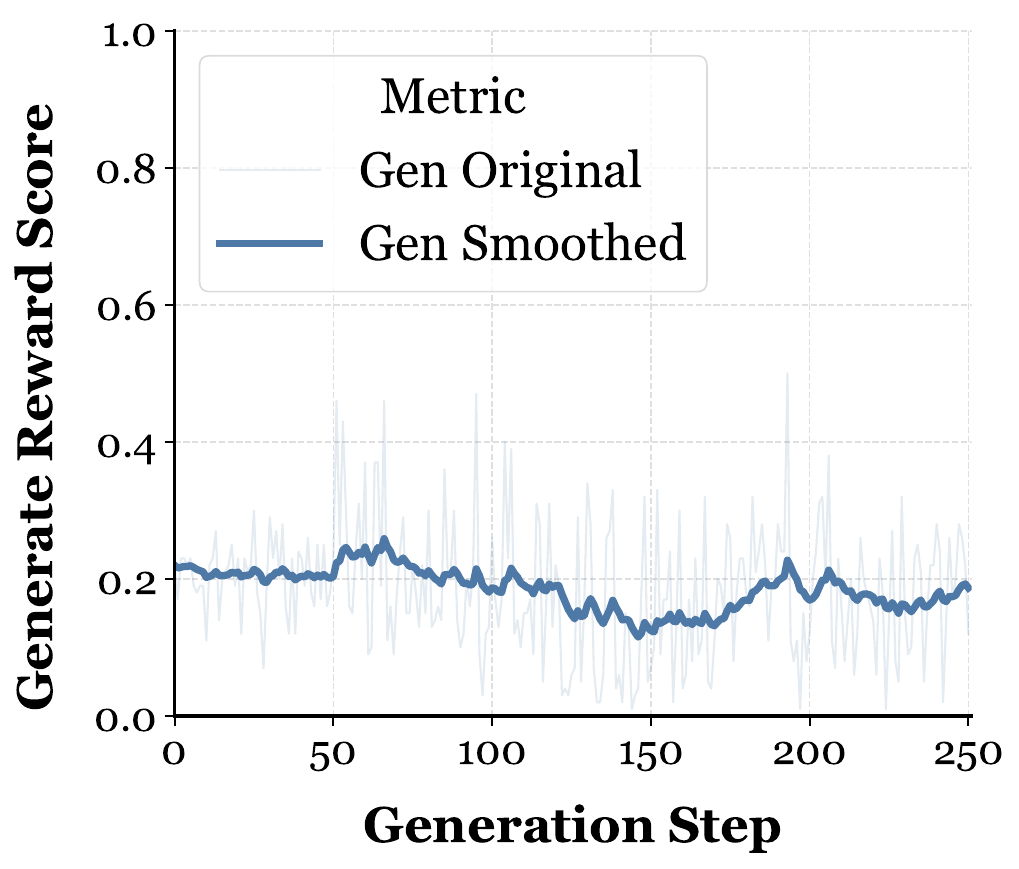}
        \vspace{-1em}
        \caption{Qwen2.5-7B}
        \label{F6_8}
    \end{subfigure}

    \caption{(a–b) Plan-level and generation-level reward convergence for LLaMA3.1-8B. (c–d) Plan-level and generation-level reward convergence for Qwen2.5-7B, demonstrating stable optimization across the two levels.}
    \label{fig:F8_convergence}
\end{figure*}

Figure~\ref{fig:F8_convergence} presents plan-level and generation-level reward traces for LLaMA3.1-8B and Qwen2.5-7B. These curves show how the recorded rewards evolve over optimization steps, characterizing the training dynamics of the two offline preference optimization stages.

\textbf{Offline Preference Optimization Enables Stable Improvement.}
The shared reward provides a consistent criterion for planning and generation. As shown in Figure~\ref{fig:F8_convergence}, reward traces characterize optimization dynamics across backbones. Table~\ref{T2} further assesses the benefits of combining both stages. This coordination occurs during offline training, while inference remains feed-forward.
% \textbf{Feedback Coupling Enables Stable Stage-wise Improvement.}
% The full LoopWeaver model shows steady gains across the four checkpoints, while models without feedback coupling exhibit fluctuations, particularly between the planning and generation checkpoints. This highlights the importance of training-time cross-stage feedback.

\subsection{Case Study: Prompt-Based Generation Across Qwen2.5 Variants (RQ6)}
\label{5.7}
We present a comparative case study using three Qwen2.5 variants of different sizes under identical prompts and constraints to generate life narratives across varying model scales and capacities.
\begin{figure*}[h]
    \centering
    % \small
    % \begin{subfigure}[b]{1.0\textwidth}
    %     \centering
    \includegraphics[width=\linewidth]{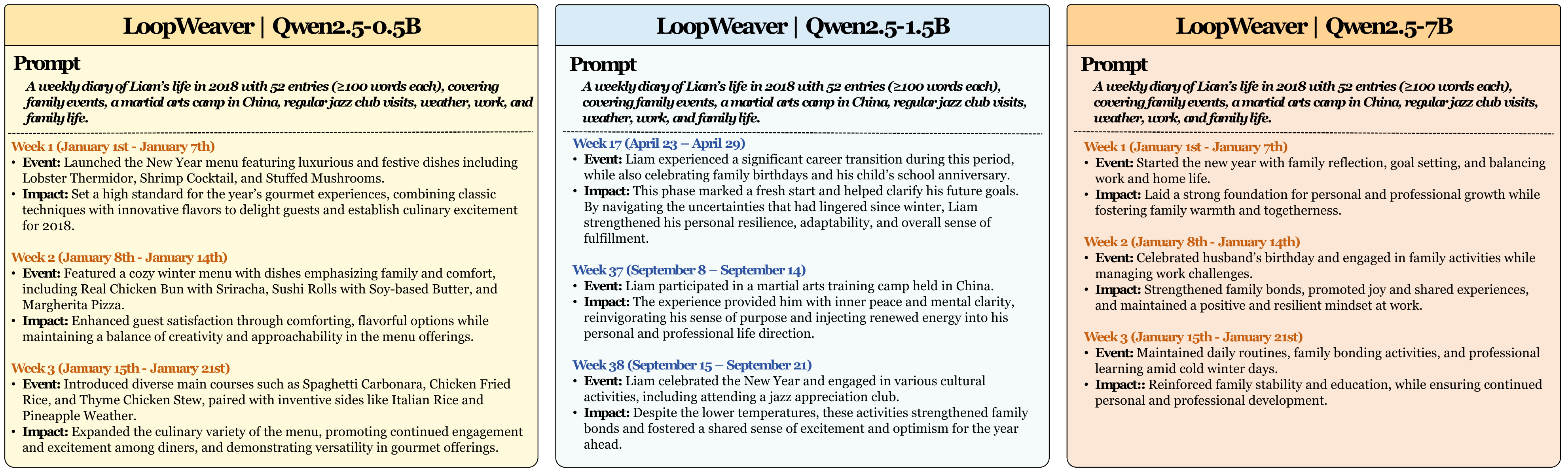}
    %     \vspace{-1.5em}
    %     \caption{Case study on generation quality under a representative constraint prompt, comparing different Qwen2.5 model variants.}
    %     \label{fig:case_study}
    % \end{subfigure}
    \caption{Case study illustrating generation quality under a representative constraint prompt.}
    % \label{fig:F6_case}
    \label{fig:case_study}
\end{figure*}

In this case study, the 7B variant shows greater thematic depth and stronger cross-paragraph consistency than the smaller variants, which more often exhibit fragmented logic, repetitive content, and weaker long-range coherence. The 7B output develops its themes through more specific details and clearer connections between events, allowing paragraphs to contribute to a coherent narrative. Its transitions provide a smoother progression of ideas, while elaboration helps explain the significance of events rather than merely listing them. In contrast, the smaller variants tend to revisit similar descriptions without advancing the narrative, and their paragraphs are less consistently connected. The 7B variant therefore stands out in both the richness of its content and the continuity of its development, sustaining a more detailed and cohesive account across the generated text.
\section{Conclusion}
In this work, we introduce LoopWeaver, a hierarchical workflow for constrained long-form text generation. By integrating \emph{constraint-aware planning}, \emph{feasibility filtering}, and \emph{reward-based preference optimization}, LoopWeaver coordinates planning and generation through two sequential offline DPO stages that share a deterministic reward. Experiments on four LongGenBench-style structured scenarios show improvements in both average text quality and overall constraint-following accuracy across all four open-source backbones. LoopWeaver achieves the highest average scores among the compared methods within each backbone group, demonstrating the benefits of combining structured planning with preference-optimized generation. Future work will improve planning reliability for smaller models and extend evaluation to long-form tasks with complex structural requirements.
% \section*{References}
\bibliographystyle{plainnat}  % NeurIPS-compatible natbib style
\bibliography{neurips}        % refs.bib 文件名（不带后缀）

% References follow the acknowledgments in the camera-ready paper. Use unnumbered first-level heading for
% the references. Any choice of citation style is acceptable as long as you are
% consistent. It is permissible to reduce the font size to \verb+small+ (9 point)
% when listing the references.
% Note that the Reference section does not count towards the page limit.
% \medskip

% {
% \small

% [1] Alexander, J.A.\ \& Mozer, M.C.\ (1995) Template-based algorithms for
% connectionist rule extraction. In G.\ Tesauro, D.S.\ Touretzky and T.K.\ Leen
% (eds.), {\it Advances in Neural Information Processing Systems 7},
% pp.\ 609--616. Cambridge, MA: MIT Press.

% [2] Bower, J.M.\ \& Beeman, D.\ (1995) {\it The Book of GENESIS: Exploring
%   Realistic Neural Models with the GEneral NEural SImulation System.}  New York:
% TELOS/Springer--Verlag.

% [3] Hasselmo, M.E., Schnell, E.\ \& Barkai, E.\ (1995) Dynamics of learning and
% recall at excitatory recurrent synapses and cholinergic modulation in rat
% hippocampal region CA3. {\it Journal of Neuroscience} {\bf 15}(7):5249-5262.
% }

%%%%%%%%%%%%%%%%%%%%%%%%%%%%%%%%%%%%%%%%%%%%%%%%%%%%%%%%%%%%
\newpage
\appendix
\onecolumn
\section{Prompts Used in LoopWeaver}
\subsection{Plan Construction Prompt}
\label{AppendixA1}

\noindent\textbf{Week Plan Prompt.}  
The Week Plan Construction Prompt guides the model to create a 52-week plan based on user requirements. It identifies special events and their exact weeks, including periodic events. The model generates a structured weekly plan in JSON format, and a revision step ensures all events strictly follow user specifications.

\begin{center}
\includegraphics[width=0.75\linewidth, keepaspectratio]{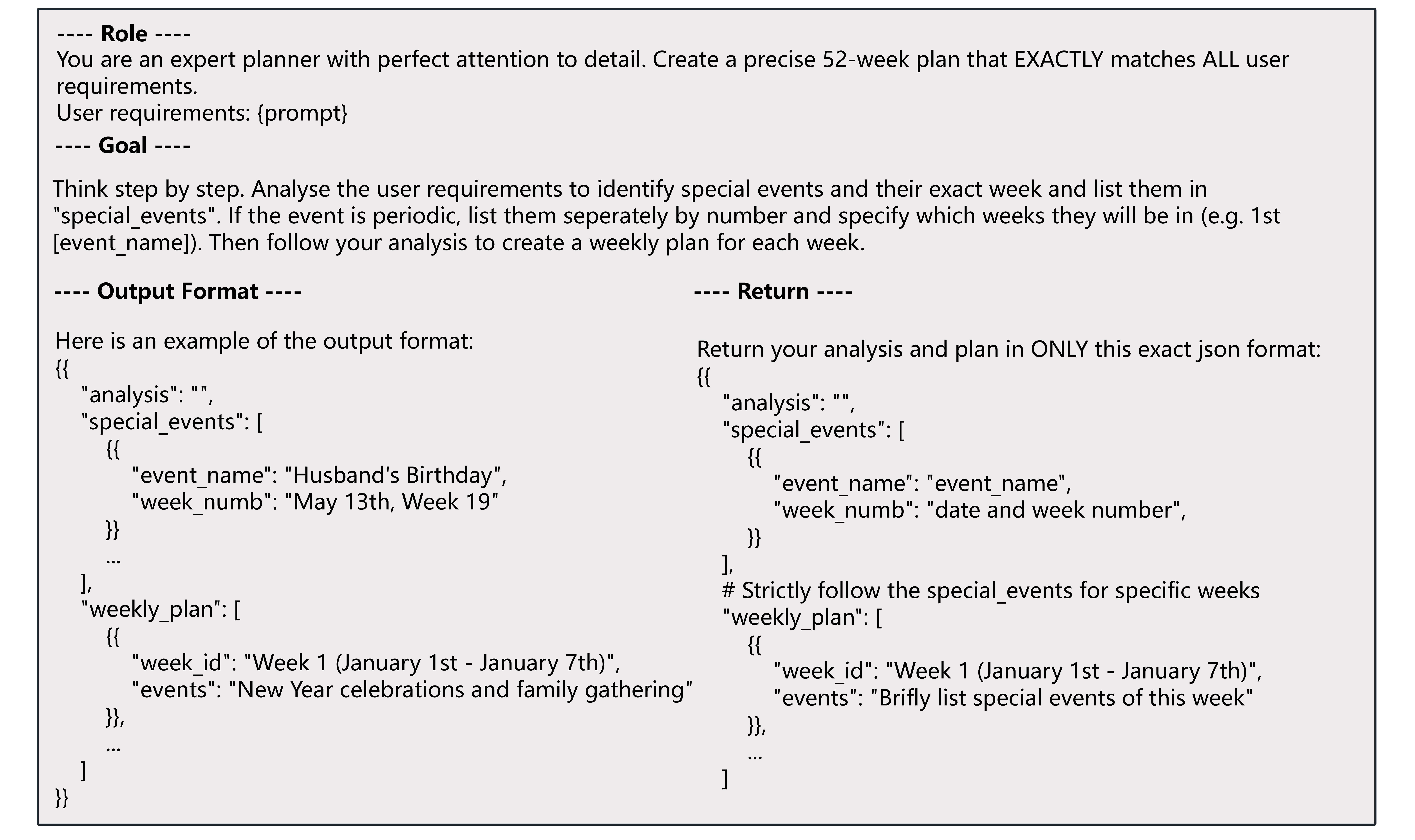}
\captionof{figure}{Plan Construction Prompt illustration}
\label{F7}
\end{center}

\noindent\textbf{Floor Plan Prompt.}  
The Floor Plan Prompt instructs the model to create a detailed floor-by-floor plan for a skyscraper. Special facilities are assigned to specific floors, including recurring ones, and the plan is output in JSON format.

\medskip
\begin{center}
\includegraphics[width=0.75\linewidth, keepaspectratio]{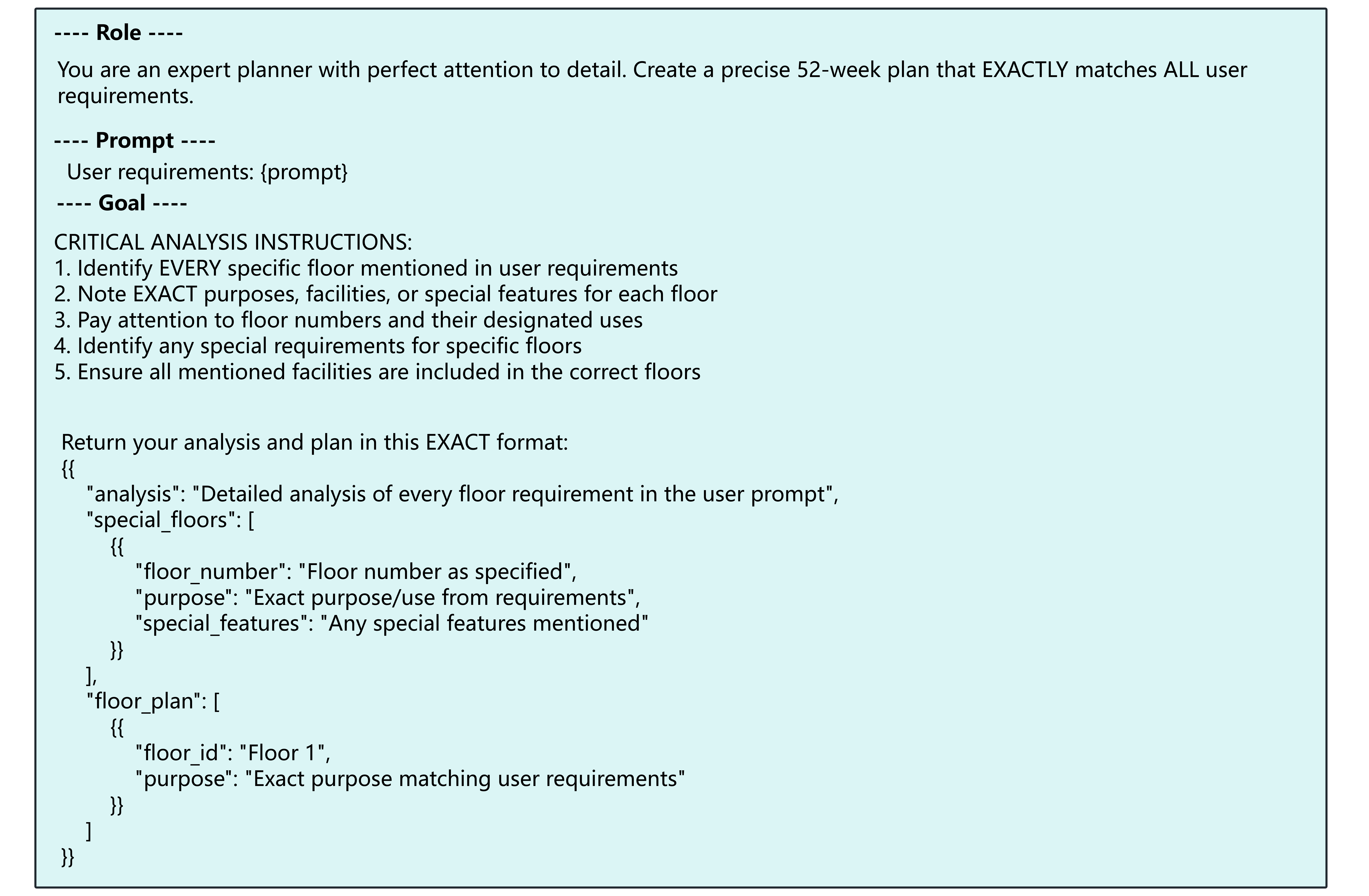}
\captionof{figure}{Floor Plan Prompt illustration}
\label{F8}
\end{center}
\medskip

\newpage
\noindent\textbf{Menu Plan Prompt.}  
The Menu Plan Prompt generates a 52-week menu schedule, assigning special dishes to specific weeks, including periodic ones.

\medskip
\begin{center}
\includegraphics[width=0.75\linewidth, keepaspectratio]{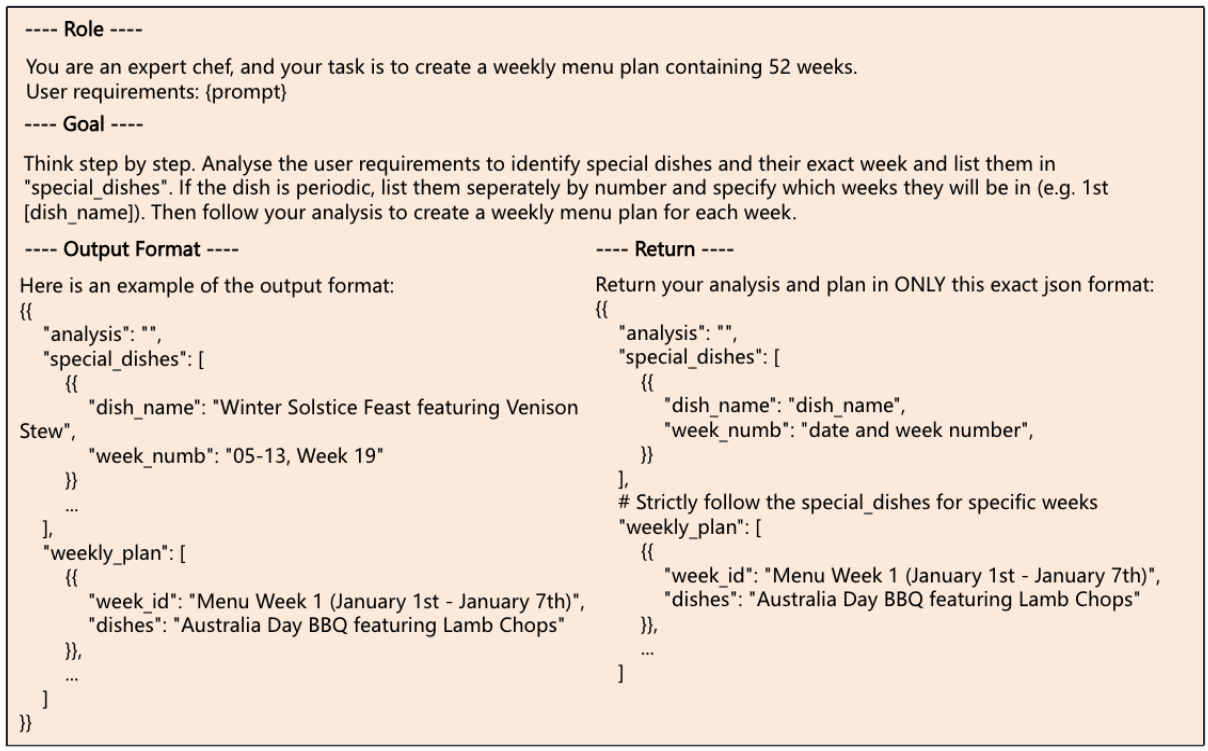}
\captionof{figure}{Menu Plan illustration}
\label{F9}
\end{center}
\medskip

\noindent\textbf{Block Plan Prompt.}  
The Block Plan Prompt guides the design of a 10×10 city block grid, assigning special uses to specific blocks and handling periodic patterns.

\medskip
\begin{center}
\includegraphics[width=0.75\linewidth, keepaspectratio]{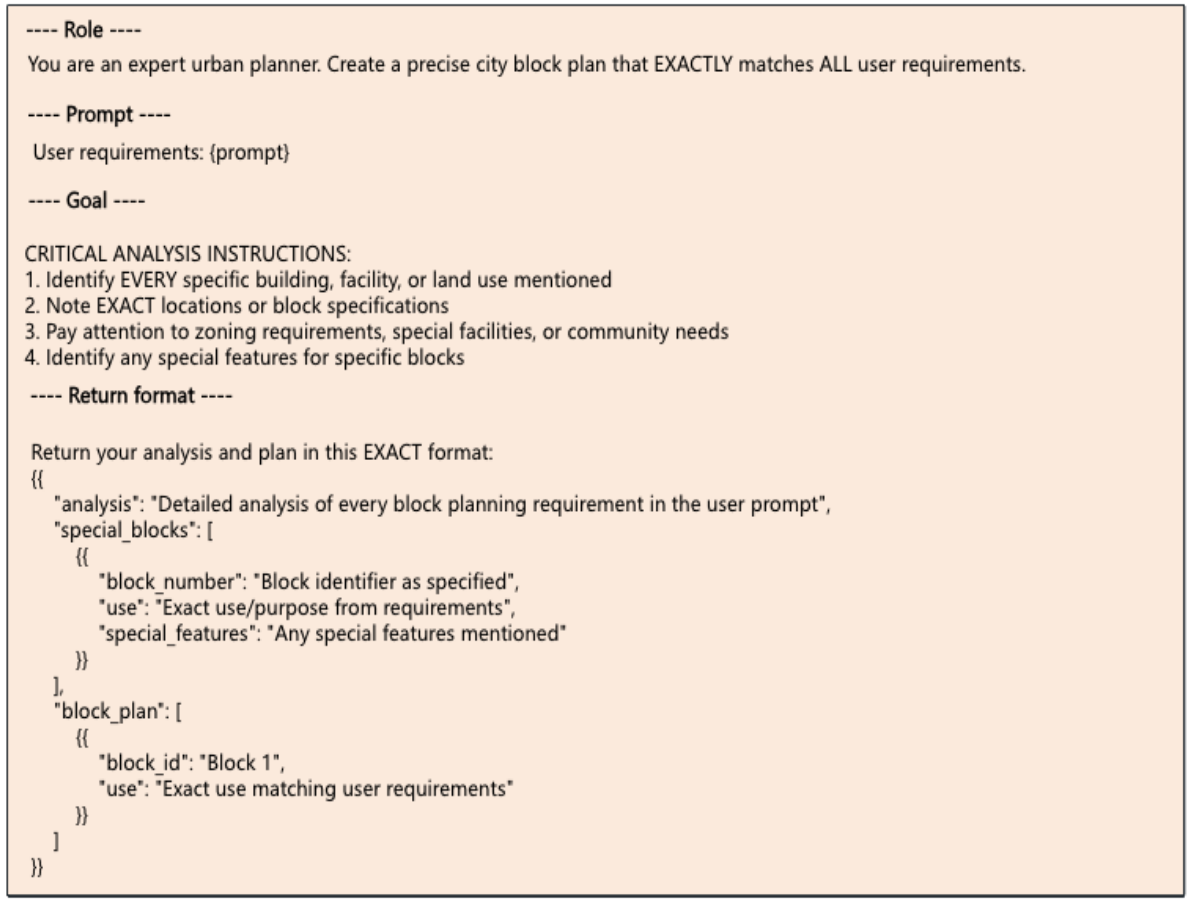}
\captionof{figure}{Block Plan Prompt illustration}
\label{F10}
\end{center}
\medskip

\newpage
\subsection{Writing Prompt}
\label{AppendixA2}

\noindent\textbf{Weekly Diary Prompt.}  
The Weekly Diary Prompt guides the model to write a 200-word diary entry for a given week. It incorporates weekly events, ensures coherence with the overall yearly plan, and checks user requirements for special events. The model outputs a structured JSON containing the week identifier, a reasoning check, and the diary entry text.  

\medskip
\begin{center}
\includegraphics[width=0.95\linewidth, keepaspectratio]{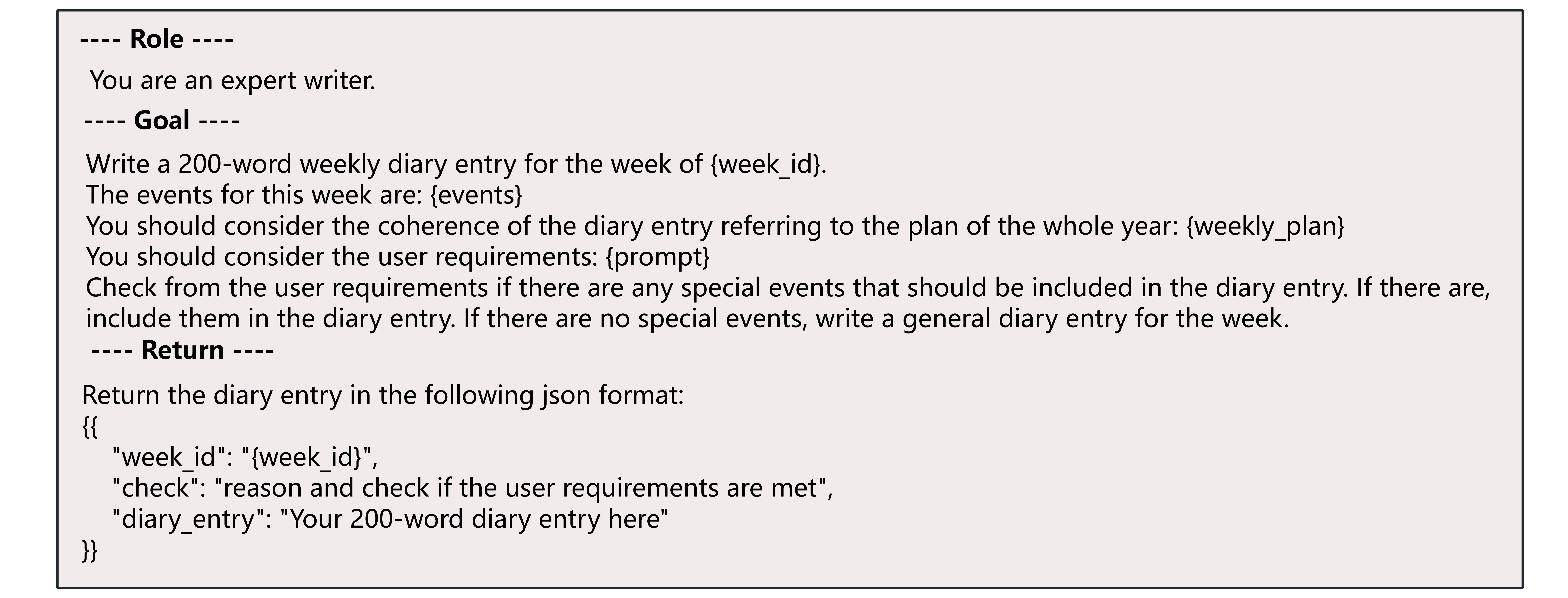}
\captionof{figure}{Weekly Diary Prompt}
\label{F11}
\end{center}
\medskip

\noindent\textbf{Skyscraper Floor Plan Prompt.}  
The Skyscraper Floor Plan Prompt guides the model to design a 150-word floor plan for a specific skyscraper floor. It considers the floor’s purpose, maintains coherence with the overall skyscraper plan, and verifies any special requirements from the user. The output is a structured JSON including the floor identifier, a reasoning check, and the generated floor plan.  

\medskip
\begin{center}
\includegraphics[width=0.95\linewidth, keepaspectratio]{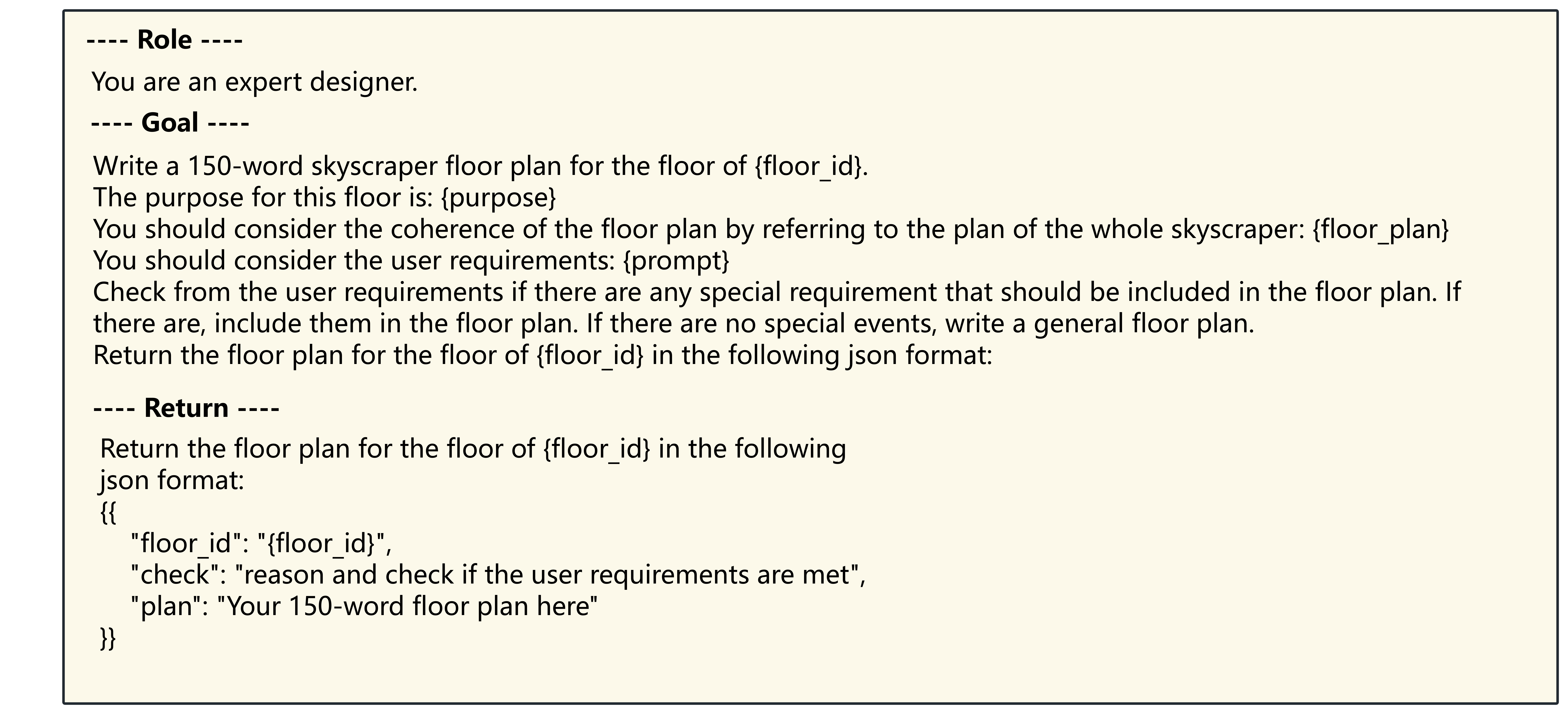}
\captionof{figure}{Skyscraper Floor Plan Prompt}
\label{F12}
\end{center}
\medskip

\newpage
\noindent\textbf{Weekly Menu Plan Prompt.}  
The Weekly Menu Plan Prompt guides the model to create a 200-word menu plan for a given week. It incorporates provided dishes, ensures coherence with the yearly menu plan, and verifies special dishes from the user requirements. The model returns a structured JSON with the week identifier, a reasoning check, and the weekly menu plan.  

\medskip
\begin{center}
\includegraphics[width=0.95\linewidth, keepaspectratio]{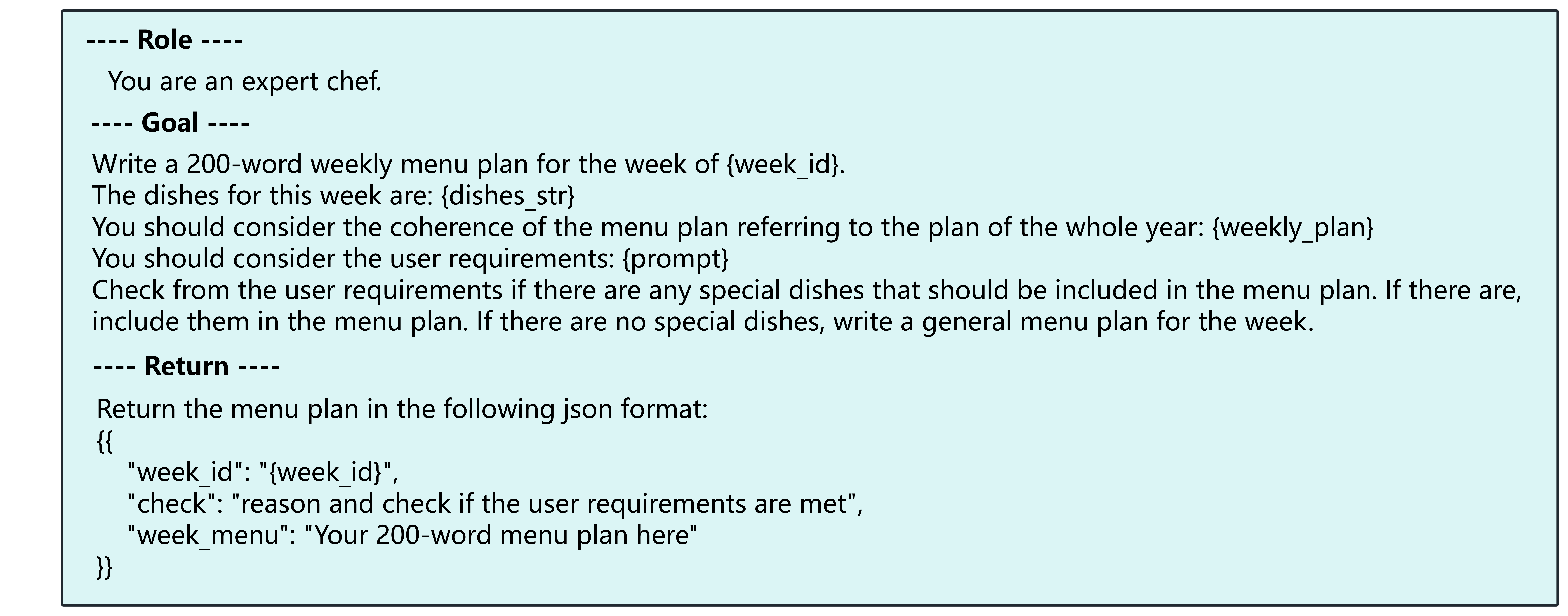}
\captionof{figure}{Weekly Menu Plan Prompt}
\label{F13}
\end{center}
\medskip

\noindent\textbf{City Block Plan Prompt.}  
The City Block Plan Prompt guides the model to design a 150-word city block plan. It considers the intended use of the block, ensures coherence with the overall city plan, and checks for any special user requirements. The generated output is returned in JSON format, including the block identifier, a reasoning check, and the block plan.  

\medskip
\begin{center}
\includegraphics[width=0.95\linewidth, keepaspectratio]{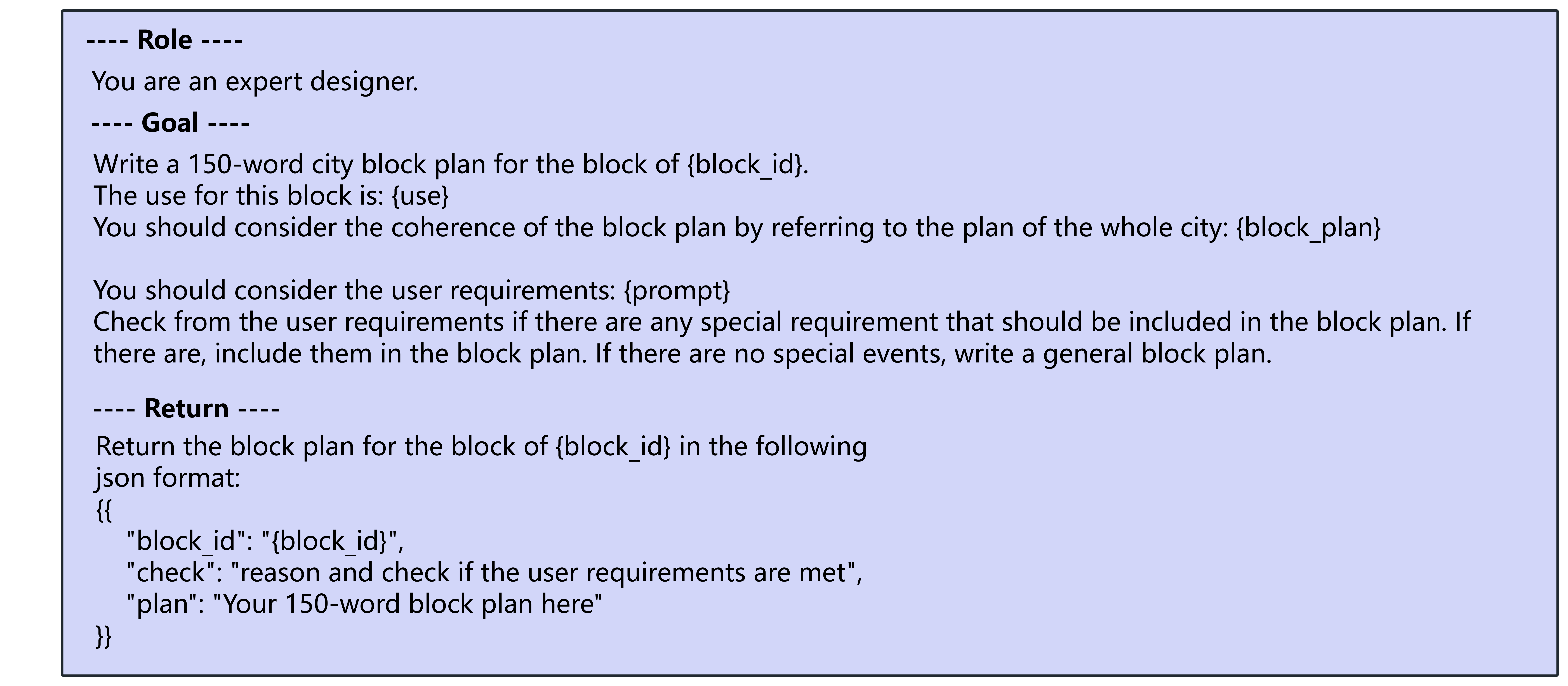}
\captionof{figure}{City Block Plan Prompt}
\label{F14}
\end{center}
\medskip

\newpage
\subsection{Binary decisions Prompt}
\label{AppendixA3}

As shown in Figure~\ref{F15}, this prompt template evaluates a plan's relevance to a given question. The judgment (YES/NO) is based on six criteria: direct relevance, completeness, logical coherence, efficiency, specificity, and consistency with the prompt context. Examples illustrate common failures—such as irrelevant steps, missing actions, or illogical order. The structured format includes the prompt, the plan, and the binary relevance label, enabling systematic assessment of planning quality.

\medskip
\begin{center}
\includegraphics[width=0.95\linewidth, keepaspectratio]{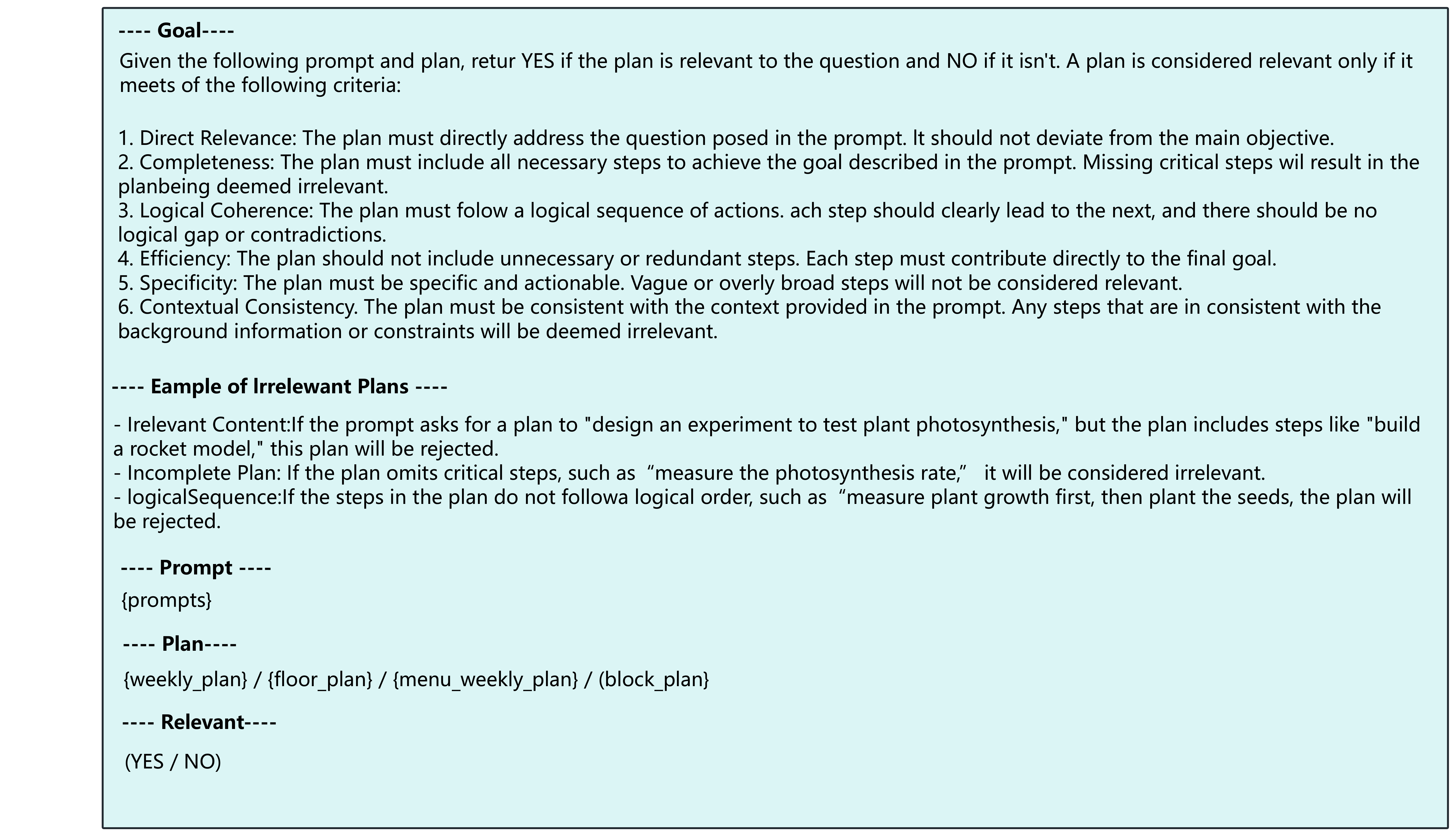}
\captionof{figure}{Prompt for binary decisions}
\label{F15}
\end{center}
\medskip

\newpage
\section{Conditional Analysis and Empirical Verification}
\label{proofs}
\subsection{Conditional analysis of planning refinement}
\label{proof1}

\textbf{Empirical Observation 1.} \textit{Constraint-aware hierarchical planning is associated with improved plan validity and coverage under the stated monotonicity assumption.}

\paragraph{Conditional analysis.}
To formalize the analysis, we introduce a mild and sufficient assumption on the
refinement operator.

\paragraph{Assumption 1 (Monotonic Improvement under Constraint-Aware Refinement).}
Let $p$ denote a hierarchical plan and $\mathcal{R}$ the refinement operator
described in Section~\ref{sec:plan-modeling}. We assume that there exists a plan quality function
$f(p) \in \mathbb{R}$, measuring constraint satisfaction and structural validity,
such that
\begin{equation}
\mathbb{E}\big[f(\mathcal{R}(p)) \mid p\big] \ge f(p).
\end{equation}
This assumption does not claim that arbitrary refinement steps improve plan
quality. Instead, it formalizes a sufficient condition satisfied by our
constraint-aware hierarchical refinement, as shown below.

Let a hierarchical plan be decomposed into sub-plans
\begin{equation}
p_i = ( S_{i,1}, S_{i,2}, \dots, S_{i,m_i} ), \quad S_{i,j} \in \mathcal{S},
\end{equation}
where each sub-plan $S_{i,j}$ is associated with the local constraint
subset $C_{i,j} \subseteq C$ inherited from the global constraint set $C$.

We define the plan quality function as an additive decomposition over sub-plans:
\begin{equation}
f(p_i)
\;=\;
\sum_{j=1}^{m_i}
f_j\!\left(S_{i,j} \mid C_{i,j}\right),
\end{equation}
where each local score $f_j(\cdot)$ is nonnegative and evaluates the constraint
satisfaction and structural validity of sub-plan $S_{i,j}$.

For the benchmark-level reporting, constraint coverage is defined as
\begin{equation}
q(p;C)=\frac{1}{|C|}\sum_{c\in C}\mathbb{I}[p\models c],
\end{equation}
where $p\models c$ indicates that plan $p$ satisfies constraint $c$. We use validity to denote the corresponding binary satisfaction of the stated structural requirements; these definitions are protocol-specific and do not constitute a formal guarantee of document-level coherence.

The following conditional lemma makes the implication explicit. Let
$\mathcal{R}(P;x,C)$ denote the complete-plan refinement operator. If, for every
constraint $c\in C$,
\begin{equation}
\mathbb{E}\!\left[\mathbb{I}[\mathcal{R}(P;x,C)\models c]\mid P,x,C\right]
\geq
\mathbb{I}[P\models c],
\end{equation}
then, by linearity of expectation,
\begin{equation}
\mathbb{E}\!\left[q(\mathcal{R}(P;x,C);C)\mid P,x,C\right]
\geq q(P;C).
\end{equation}
This is a sufficient conditional assumption for non-decreasing expected coverage,
not a universal guarantee that every refinement improves every plan.

At each refinement step, the hierarchical planner identifies a violating
sub-plan $S_{i,k}$ and applies a localized revision within the complete-plan
context, producing an updated sub-plan $S'_{i,k}$ while preserving the other
sub-plans for the consistency check:
\begin{equation}
\begin{aligned}
p_i'
= (
& S_{i,1}, \dots, S_{i,k-1}, S'_{i,k}, \\
& S_{i,k+1}, \dots, S_{i,m_i} ).
\end{aligned}
\end{equation}

Consequently, the change in plan quality depends only on the revised sub-plan:
\begin{equation}
\begin{aligned}
& \mathbb{E}\!\left[f(p_i') \mid p_i\right] - f(p_i)\\
& = \mathbb{E}\!\left[f_k(S'_{i,k}) \mid S_{i,k}\right] - f_k(S_{i,k}).
\end{aligned}
\end{equation}

Since the refinement operator $\mathcal{R}$ is constraint-aware and employs
validity filtering and top-$K$ selection (Section~\ref{sec:plan-modeling}), refinements that reduce
local validity or violate constraints are rejected with high probability.
Therefore,
\begin{equation}
\mathbb{E}\!\left[f_k(S'_{i,k}) \mid S_{i,k}\right] \ge f_k(S_{i,k}),
\end{equation}
which implies
\begin{equation}
\mathbb{E}\!\left[f(p_i') \mid p_i\right] \ge f(p_i).
\end{equation}

By iteratively applying hierarchical refinement, the expected plan quality is
non-decreasing under Assumption 1. This conditional statement does not guarantee
that every candidate plan satisfies all constraints or structural requirements.

We emphasize that this argument relies critically on hierarchical decomposition
and localized refinement. For flat planning or unconstrained revision processes,
modifying one component may arbitrarily affect others, and the above analysis no
longer holds.
\paragraph{Conclusion.} Under these assumptions, the expected plan quality is non-decreasing after refinement.

\subsection{Conditional analysis of feasibility filtering}
\label{proof2}

\textbf{Empirical Observation 2.} \textit{Feasibility filtering is associated with fewer misaligned plans passed to generation under the stated rejective-power assumption.}

\paragraph{Conditional analysis.}
Let $P_{\text{final}} = \{p_1, \dots, p_N\}$ be the set of candidate plans after
refinement. Each plan $p_i$ is evaluated by a template-driven, feasibility-oriented
screening mechanism and assigned a binary acceptance indicator
\begin{equation}
\delta_i = \mathbf{1}_{\mathrm{relevant}}(p_i) \in \{0,1\},
\end{equation}
where $\delta_i = 1$ indicates that the plan satisfies the task-inherent relevance
criteria defined in Section~\ref{sec:binary-filter}, and $\delta_i = 0$ otherwise. The filtered plan
set is given by
\begin{equation}
P_{\text{filtered}} = \{ p_i \in P_{\text{final}} \mid \delta_i = 1 \}.
\end{equation}

Let $\phi(p) \ge 0$ denote a nonnegative violation score measuring the degree to
which plan $p$ violates task constraints or semantic requirements, with
$\phi(p)=0$ indicating full validity.

We do not assume that the plan generator produces valid plans with high
probability, nor that the screening mechanism perfectly separates high- and
low-quality plans. Instead, we impose a minimal and realistic assumption on the
feasibility-oriented screening process.

\paragraph{Assumption 2 (Non-trivial Rejective Power).}
There exist constants $\tau > 0$ and $\eta > 0$ such that for any plan $p$ with
$\phi(p) \ge \tau$,
\begin{equation}
\Pr(\delta = 0 \mid p) \ge \eta.
\end{equation}
That is, plans exhibiting sufficiently large violations are rejected with
non-zero probability. Here, the probability notation abstracts over the
stochasticity introduced by the template-driven LLM evaluator and decoding
process, rather than assuming a calibrated probabilistic classifier.

This assumption is weak and does not require perfect discrimination; it only
formalizes that the screening mechanism has some ability to reject clearly
invalid plans.

Under this assumption, the probability that a retained plan exhibits a large
violation is bounded as
\begin{equation}
\Pr\big( \phi(p) \ge \tau \mid \delta = 1 \big)
\le
\frac{\Pr\big( \phi(p) \ge \tau \big) (1-\eta)}
{\Pr(\delta = 1)}.
\end{equation}

The bound above provides an upper bound on the violation risk among retained
plans. If the additional condition
$\Pr(\delta=1)>1-\eta$ holds, this bound is strictly lower than the
unfiltered violation probability. Without this additional condition, the
analysis does not imply a strict reduction, but it still characterizes the
risk induced by feasibility-oriented screening. Thus, the filter can reduce
the risk of propagating highly invalid or semantically misaligned plans to
downstream generation under the stated condition.

\subsection{Empirical verification of reward-based preference optimization}
\label{proof3}
\textbf{Empirical Observation 3.} \textit{Reward-based preference optimization aligns plans and segments with task objectives.}

\paragraph{Empirical verification.}
After binary relevance filtering, each plan is decomposed into sub-plans for optimization, followed by content generation guided by the sub-plan. For each candidate \( X \) (a sub-plan \( S_j \) or a generated segment \( G_k \)), let $c(X)$ denote its conditioning context: $c(S_j) = (x,C)$ at the plan level and $c(G_k) = (x, C, S_k)$ at the generation level. We define its reward by direct deterministic evaluation:
\begin{equation}
r(X) = R\bigl(X,\, (x,C)\bigr),
\end{equation}
where $R$ is the shared rule-based reward defined in Section~\ref{sec:preliminaries}.

Recall the DPO objective used in the paper:

\begin{equation}
\begin{aligned}
\Delta(X^+, X^-) 
&= \beta \log \frac{\pi_\theta(X^+ \mid c(X))}{\pi_{\text{ref}}(X^+ \mid c(X))} \\
&\quad - \beta \log \frac{\pi_\theta(X^- \mid c(X))}{\pi_{\text{ref}}(X^- \mid c(X))}, \\[1ex]
\mathcal{L}_{\text{DPO}}(\theta) 
&= - \mathbb{E}_{(X^+, X^-) \sim D} \\ 
& \Big[ 
\log \sigma \big( \Delta(X^+, X^-) \big) 
\Big].
\end{aligned}
\end{equation}
This objective encourages the model policy $\pi_\theta$ to increase the relative log-probability ratio for higher-reward candidates $(X^+)$ and decrease it for lower-reward candidates $(X^-)$, proportionally to the preference signal induced by the reward difference.

In the implemented training procedure, preference labels are binary:
$X^+$ is selected when $r(X^+)>r(X^-)$, while $X^-$ is the lower-reward
candidate. No soft reward-probability label is used. The DPO objective
therefore increases the relative likelihood of the higher-reward candidate
under the shared conditioning context $c(X)$. The resulting alignment is
evaluated empirically in the main text.

Because preference pairs are ranked directly by the deterministic reward, the DPO objective shifts model likelihood toward candidates with higher constraint coverage and prompt relevance. The resulting alignment is evaluated empirically in the main text.
\paragraph{Conclusion.} The analysis describes the direction of the DPO preference signal; quantitative gains are established by the experiments in the main text.

\newpage
\section{LoopWeaver Algorithm Details}
Algorithm~\ref{alg:LoopWeaver} realizes the three components of Section~\ref{sec:methodology} together with the offline DPO training that couples the two levels: (1)~\emph{Constraint-aware hierarchical planning} (Section~\ref{sec:plan-modeling}) generates a complete plan from the writing prompt and constraint set $C$, revises sub-plans in full-plan context, and retains the resulting candidates. (2)~\emph{Feasibility filtering} (Section~\ref{sec:binary-filter}) applies one prompt-only YES/NO decision to each complete plan using the six task-inherent criteria. (3)~\emph{Reward-based preference optimization} (Section~\ref{sec:reward-opt}) constructs preference pairs at both plan and generation levels using the deterministic reward $R(\cdot, e)$ defined in Section~\ref{sec:preliminaries}, and updates the policy via the DPO objectives $\mathcal{L}^{\text{plan}}_{\text{DPO}}$ and $\mathcal{L}^{\text{gen}}_{\text{DPO}}$. The shared reward couples the two levels during offline training; inference is feed-forward and incurs no live feedback overhead.

\begin{algorithm}[h]
\small
\caption{LoopWeaver: Hierarchical Offline Preference Optimization for Constrained Long-Form Text Generation}
\label{alg:LoopWeaver}
\begin{algorithmic}[1]
\Require Prompt $x$, structured constraint set $C = C_{\text{once}} \cup C_{\text{range}} \cup C_{\text{periodic}}$, policy $\pi_\theta$, reference policy $\pi_{\text{ref}}$, deterministic rule-based reward $R(\cdot, e)$ with $e=(x, C)$ as defined in Section~\ref{sec:preliminaries}~(ii), candidate plan count $K$, DPO temperature $\beta$
\Ensure Updated policy $\pi_\theta$

\vspace{1mm}
\State \textbf{// Component 1: Constraint-Aware Plan Modeling (Section~\ref{sec:plan-modeling})}
\For{$i = 1$ to $K$}
    \State Sample complete candidate plan $p_i \sim \pi_\theta(\cdot \mid x, C)$
    \State Revise each sub-plan in full-plan context: $p_i \gets \mathcal{R}(p_i; x, C)$ \Comment{Eq.~\ref{eq:local-repair}}
\EndFor
\State $P_{\text{final}} \gets \{p_1, \dots, p_K\}$

\vspace{1mm}
\State \textbf{// Component 2: Binary Relevance Filtering (Section~\ref{sec:binary-filter})}
\For{each $p_i \in P_{\text{final}}$}
    \State Query the same backbone with the prompt-only YES/NO filter
    \State Set $\delta_i = \mathbf{1}[\,p_{\text{filter}}(p_i,x,C)=\text{YES}\,]$
\EndFor
\State $P_{\text{filtered}} \gets \{p_i \in P_{\text{final}} \mid \delta_i = 1\}$

\vspace{1mm}
\State \textbf{// Component 3: Reward-Guided Optimization (Section~\ref{sec:reward-opt})}
\State \textit{Plan-level DPO:}
\For{each plan $p \in P_{\text{filtered}}$ and each sub-plan $S_j \in p$}
    \State Compute deterministic unit reward $r(S_j) = R(S_j,e)$ \Comment{Eq.~\ref{eq:plan-reward}}
\EndFor
\State Construct preference pairs $(S_j^+, S_j^-)$ with $r(S_j^+) > r(S_j^-)$ to form $D_{\text{plan}}$
\State Update $\pi_\theta$ by minimizing $\mathcal{L}^{\text{plan}}_{\text{DPO}}(\theta)$ over $D_{\text{plan}}$

\vspace{0.5mm}
\State \textit{Generation-level DPO:}
\For{each sub-plan $S_k$ from updated $\pi_\theta$}
    \State Sample candidate segments $G_k \sim \pi_\theta(\cdot \mid x, S_k)$
    \State Compute deterministic unit reward $r(G_k) = R(G_k,e)$
\EndFor
\State Construct preference pairs $(G_k^+, G_k^-)$ with $r(G_k^+) > r(G_k^-)$ to form $D_{\text{gen}}$
\State Update $\pi_\theta$ by minimizing $\mathcal{L}^{\text{gen}}_{\text{DPO}}(\theta)$ over $D_{\text{gen}}$

\State \Return updated policy $\pi_\theta$
\end{algorithmic}
\end{algorithm}

\newpage
\section{Dataset Details}
\label{Dataset}

Our dataset is constructed based on the official LongGenBench codebase, following the methodology described in Section~\ref{sec:methodology} of the main paper. It comprises \textbf{400 synthetic long-form planning and generation tasks}, spanning diverse domains such as weekly event planning, architectural floor designs, menu organization, and city block descriptions. Each task instance is automatically paired with structured prompts, hierarchical global plans, task-specific constraints (once, range, or periodic), and evaluation prompts, enabling the dataset to jointly capture global planning information and local generation requirements. Each sample includes \textbf{(i)} a global hierarchical plan, \textbf{(ii)} embedded structural constraints, \textbf{(iii)} sub-plan prompts used for preference optimization, and \textbf{(iv)} the final generation targets. On average, global plans contain 250--300 words, while individual generated segments contain 150--200 words depending on the task type.

Although the dataset is synthetically generated, this construction is both appropriate and necessary for our benchmark. Real-world human-authored corpora rarely contain explicit structural constraints—such as hierarchical outlines, length-range specifications, and periodic content formats—at the scale and diversity required for systematic evaluation. Synthetic generation enables precise and reproducible control over constraint complexity, ensuring consistent coverage across task categories without compromising the realism of structural patterns. Importantly, the dataset serves as a \textit{task specification} rather than a corpus for style imitation: LoopWeaver learns to follow constraints, not to mimic LLM-produced writing styles. Since the model never encounters test instances during training and DPO relies solely on preference signals rather than memorizing dataset patterns, the synthetic origin does not introduce data leakage. Moreover, the structural constraints modeled in this dataset—such as multi-level sectioning, fixed-length segments, and periodic organization—naturally occur in real-world long-form writing, including reports, educational documents, and technical manuals, making the resulting capabilities transferable to human-authored scenarios.

The 400 benchmark samples described above are reserved for held-out evaluation when comparing different model backbones (Qwen2.5, GPT-4o-mini, and LLaMA3.1) as reported in Section~\ref{sec:experiments}. DPO preference pairs are generated from a separate synthetic training split using the same task templates but different prompts and seeds; none of the 400 evaluation prompts or outputs is used in DPO training. To ensure reproducibility, we adopt identical sampling parameters as the official LongGenBench implementation. Since all samples are generated from synthetic prompts and LLM-produced outputs, the dataset contains no personally identifiable information (PII) or sensitive content and is intended solely for academic research on long-form planning and generation.

\section{Baseline Details}
\label{Baseline}

\textbf{GPT-4o-mini.} In our experiments, we evaluate the proposed \textbf{LoopWeaver} framework against \textbf{CogWriter}, a multi-step cognitive planning and generation framework specifically designed for long-form text. CogWriter performs hierarchical planning to structure content and leverages GPT-4o-mini for each generation step, allowing it to handle complex long-form tasks with temporal and spatial constraints.

\textbf{Qwen2.5.} For models based on {Qwen2.5} across multiple scales (0.5B, 1.5B, and 7B), we also compare against both \textbf{CogWriter} and \textbf{LongWriter}. In this setting, CogWriter utilizes Qwen2.5 as the generation backbone, employing hierarchical planning and sub-plan guided generation to produce coherent outputs for long-form scenarios such as diary writing, menu design, skyscraper floor planning, and urban block layouts. LongWriter, by contrast, relies on extended-context training and direct generation without explicit planning. Including these baselines across different model capacities enables a systematic evaluation of how LoopWeaver improves over strong planning-augmented and data-centric long-form generation methods under varying model scales.

\textbf{LLaMA3.1-8B.} To provide a cross-backbone comparison, we include experiments based on {LLaMA3.1-8B}. For this backbone, we compare LoopWeaver against \textbf{CogWriter} and \textbf{LongWriter} as representative long-form generation baselines, as well as the base LLaMA3.1-8B model without workflow augmentation. This setting tests whether the workflow-level planning and reward-guided optimization transfer across backbones within the same structured benchmark; it is not evidence of unrestricted cross-domain generalization.

\newpage
\section{Evaluation Details}
\label{Evaluation}

For each constraint type, we compute accuracy as the fraction of prompts for which the model output satisfies the constraint:

\textbf{(a) Once constraints ($C_{\text{once}}$):} checks adherence to individual constraints.
\begin{equation}
\text{Acc}_{\text{once}} = \frac{\sum_{i=1}^{N_\text{once}} R_i}{N_\text{once}}
\end{equation}
\textbf{(b) Range constraints ($C_{\text{range}}$):} checks adherence to constraints that span multiple blocks or a specified range.
\begin{equation}
\text{Acc}_{\text{range}} = \frac{\sum_{i=1}^{N_\text{range}} R_i}{N_\text{range}}
\end{equation}
\textbf{(c) Periodic constraints ($C_{\text{periodic}}$):} checks adherence to recurring constraints across periodic intervals.
\begin{equation}
\text{Acc}_{\text{periodic}} = \frac{\sum_{i=1}^{N_\text{periodic}} R_i}{N_\text{periodic}}
\end{equation}
Here, \(R_i = 1\) if the model’s output satisfies the corresponding constraint and \(R_i = 0\) otherwise; \(N_\text{once}, N_\text{range}, N_\text{periodic}\) denote the total number of prompts for each constraint type.

\textbf{Average Accuracy (Avg. Acc)}
The \textbf{Average Accuracy} is defined as the mean of the three type-specific accuracies:
\begin{equation}
\text{Avg. Acc} = \frac{\text{Acc}_{\text{once}} + \text{Acc}_{\text{range}} + \text{Acc}_{\text{periodic}}}{3}
\end{equation}
This provides an overall measure of the model’s adherence to all constraint types.

\paragraph{Text Quality Evaluation.}
Each judge request contains the complete original prompt and
the complete generated response. The judge assigns integer scores
from 1 to 5 for Narrative Continuity, Entity \& Memory Persistence,
Temporal Groundedness, and Affective Consistency.
For method $m$, let $\mathcal{I}_m$ denote the evaluated prompt set.
The score for dimension $d$ in run $r$ is
\begin{equation}
Q_{m,d}^{(r)}
=
\frac{1}{|\mathcal{I}_m|}
\sum_{i\in\mathcal{I}_m}
25\bigl(s_{m,i,d}^{(r)}-1\bigr).
\end{equation}
The overall score is the arithmetic mean of the four dimension
scores. 

\paragraph{Repeated Evaluation.}
We perform three independent generation runs on the same test prompts and evaluate each run using the same protocol. We report the mean and sample standard deviation across runs, with the overall standard deviation computed from the three overall scores.

\newpage
\section{Implementation Details}
\label{Implementation}

All baselines and LoopWeaver are implemented based on the LongGenBench framework. Experiments are conducted on \textbf{3 NVIDIA A40 GPUs (48 GB each)}, with typical runtime reported for different model scales (0.5B / 1.5B / 7B) and average GPU memory usage during training and inference.

\subsection{Reproducibility Settings}
\label{Reproducibility}

For full reproducibility we list below the exact LoRA configuration, DPO training hyperparameters, and dataset construction protocol used in all experiments.

\paragraph{LoRA Settings.} Both the planning-stage and generation-stage DPO use the same LoRA configuration (Table~\ref{tab:lora-settings}). After each DPO stage finishes, the LoRA weights are merged back into the base model before proceeding to the next stage, so downstream training/inference operates on a fully merged checkpoint rather than on a stack of adapters.

\begin{table}[h]
\centering
\small
\caption{LoRA configuration used for the planning and generation DPO stages.}
\label{tab:lora-settings}
\begin{tabular}{lcc}
\toprule
\textbf{Setting} & \textbf{Planning DPO} & \textbf{Generation DPO} \\
\midrule
LoRA rank ($r$)  & 16    & 16 \\
LoRA alpha       & 32    & 32 \\
LoRA dropout     & 0.05  & 0.05 \\
\bottomrule
\end{tabular}
\end{table}

\paragraph{DPO Training Hyperparameters.} The same hyperparameter set (Table~\ref{tab:dpo-hparams}) is used at both DPO stages.

\begin{table}[h]
\centering
\small
\caption{DPO training hyperparameters (shared across planning and generation stages).}
\label{tab:dpo-hparams}
\begin{tabular}{lc}
\toprule
\textbf{Hyperparameter} & \textbf{Value} \\
\midrule
Optimizer                  & AdamW \\
Learning rate              & $5 \times 10^{-6}$ \\
DPO $\beta$                & 0.1 \\
Label smoothing            & 0.05 \\
Epochs                     & 3 \\
Batch size / Grad.\ accum. & 1 / 16 \\
Precision                  & bfloat16 \\
Warmup ratio               & 0.1 \\
\bottomrule
\end{tabular}
\end{table}

\paragraph{Dataset Construction.} The benchmark contains 100 samples per task type across the four scenarios (Weekly Diary, Floor Plan, Menu Plan, Block Plan), totaling 400 samples. All sampling is seeded with \texttt{seed=42}. Training and test sets are generated independently from the synthetic-data pipeline of Appendix~\ref{AppendixA1} so that there is no overlap between the two splits, ruling out data leakage.

\subsection{Computational Overhead Analysis}
\label{Overhead}

We provide a detailed accounting of LoopWeaver's inference cost to clarify the sources of overhead and demonstrate scalability.

\paragraph{LoopWeaver vs.\ Base Model.} The latency difference between LoopWeaver and the base model stems from a \emph{fundamental difference in generation strategy}, not from the offline preference-training steps. The base model emits the entire document in a single LLM call, whereas LoopWeaver decomposes the document into $\sim$52 sub-plans and generates each segment in a separate call, conditioning on the current sub-plan and the running narrative. This segment-by-segment strategy is the prerequisite for fine-grained constraint control but naturally incurs more total inference time. CogWriter also adopts segment-by-segment generation and exhibits a similar increase in time (Figure~\ref{fig:rq4}), confirming that this overhead is intrinsic to structured generation rather than specific to LoopWeaver.

\paragraph{LoopWeaver vs.\ CogWriter.} Both methods adopt segment-by-segment generation, so their generation-phase costs ($\sim$52 LLM calls) are approximately equivalent. LoopWeaver additionally performs offline preference-data construction and two DPO stages; these costs should be reported separately from concurrent inference latency.

\begin{table}[h]
\centering
\small
\setlength{\tabcolsep}{3pt}
\caption{Audited phase-wise cost for one logged run. The benchmark used three NVIDIA A40 GPUs and concurrency 100 for the concurrent inference run.}
\label{tab:cost-audit}
\begin{tabular}{@{}p{0.33\linewidth}p{0.28\linewidth}r@{}}
\toprule
Phase & Scope/configuration & Wall-clock \\
\midrule
Planning + filtering A/B & 400 / 400 samples & 8:01:45 / 8:09:21 \\
Segment construction A/B & 251 / 241 accepted plans samples & 1:33:44 / 1:34:32 \\
Planning DPO & 220 preference pairs & 1,152 s \\
Generation DPO & 211 preference pairs & 208 s \\
Concurrent inference & 400 samples, concurrency 100 & 3:09:54 \\
\bottomrule
\end{tabular}
\end{table}

The two DPO stages total 0.38 single-GPU hours per run, or 1.13 GPU-hours across the three independent result repetitions. The concurrent run corresponds to 28.49 s/sample amortized throughput, not individual request latency. The logged run metadata preserve the hardware and concurrency settings but not per-stage maximum-token caps; we therefore do not infer or claim a token-budget comparison from these logs.

\paragraph{Per-Task Inference Breakdown.}
As shown in Figure~7(b), LoopWeaver takes approximately 815\,s per task, with the following breakdown:
\begin{itemize}[leftmargin=*]
    \item \textit{Planning} (1 LLM call): approximately 20\,s.
    \item \textit{Filtering} (1 LLM call): approximately 20\,s.
    \item \textit{Generation} ($\sim$52 LLM calls): approximately 775\,s, accounting for 95.1\% of the total inference time.
\end{itemize}
Planning and filtering together account for approximately 4.9\% of the per-task inference time. The dominant cost is segment-by-segment generation. These per-task measurements are reported separately from the concurrent benchmark in Table~6; per-task completion time and concurrent wall-clock time are not interchangeable.

\paragraph{Scalability.} LoopWeaver uses a constant number of planning and filtering calls per task, but the token-level cost of those calls grows with prompt, constraint, and document length. The generation cost scales linearly with the number of sub-plans, as in other segment-by-segment generation methods. The cross-stage coupling between planning and generation is realized through DPO training: deterministic unit-level rewards from one stage shape the preference data optimized at the other during offline training. The trained policies are then executed without a live feedback channel, so this training-time coupling incurs no additional cost at inference time. This decoupling preserves the reach of shared-reward optimization while bounding deployment cost by segment-by-segment generation alone, keeping LoopWeaver practically deployable at scale.

\section{Limitations and Future Work}
\label{limitations}

While LoopWeaver demonstrates strong performance in constrained long-form text generation, it is not without limitations. First, the segment-by-segment generation strategy required for fine-grained constraint control increases per-task inference time relative to single-pass generation. The audited logs show that offline preference-data construction dominates the reported training cost, while inference still includes the repeated segment-generation calls required by structured generation; real-time applications demanding sub-second latency may find this trade-off costly. Second, although LoopWeaver effectively handles verifiable positional and lexical constraints, extremely complex tasks with highly interdependent constraints may still challenge the planning and filtering modules, occasionally leading to sub-optimal sub-plan selection. The method does not provide a formal guarantee of document-level coherence, and generated segments are not returned to the planner during inference. Rejected plans are currently discarded or regenerated; because each prompt has only one plan, a rejected plan generally lacks an accepted same-prompt full-plan counterpart, and malformed plans may not align to the same local unit for DPO negatives. We did not run a separate parseable-rejected-plan, same-prompt-positive experiment in this audit, so the effect of retaining such negatives remains unquantified. The logged records also omit per-stage token caps, limiting direct token-budget comparisons. Third, our current evaluation focuses on four LongGenBench-style structured scenarios; these results do not establish unrestricted cross-domain generalization.

For future work, we aim to explore strategies to reduce computational cost without sacrificing output quality, such as model distillation, parallelized sub-plan generation, or more efficient reward evaluation mechanisms. Additionally, integrating more advanced constraint representation and reasoning techniques could further improve performance on highly interdependent or domain-specific tasks. Finally, extending LoopWeaver to multi-modal long-form generation and to settings with online user feedback during generation may broaden its applicability and provide more flexible solutions for real-world content creation.

%%%%%%%%%%%%%%%%%%%%%%%%%%%%%%%%%%%%%%%%%%%%%%%%%%%%%%%%%%%%
%%%%%%%%%%%%%%%%%%%%%%%%%%%%%%%%%%%%%%%%%%%%%%%%%%%%%%%%%%%%
\newpage

\section{Case Studies}
\label{appendix:case-studies}

This appendix complements the main-text case study (Section~\ref{5.7}) with three end-to-end agent traces showing how LoopWeaver decomposes a long-form constrained writing prompt into the planning, filtering, and reward-based optimization stages of Section~\ref{sec:methodology}. The three cases span different domains, namely a personal weekly diary (\S\ref{case:olivia}), a 100-floor architectural document (\S\ref{case:skyscraper}), and a chef's annual menu (\S\ref{case:chef}), and exercise all three constraint types $C_{\text{once}},\,C_{\text{range}},\,C_{\text{periodic}}$. Each case is rendered as a uniform workflow card with the fields \textsc{Scenario / Goal}, \textsc{Input}, \textsc{Agent Reasoning Steps}, \textsc{Tool / Action}, \textsc{Output}, and \textsc{Result / Benefit}, and the reasoning steps are drawn as a left-to-right pipeline so the agent's behavior can be read at a glance.

\definecolor{cardblue}{RGB}{225, 236, 255}
\definecolor{cardgreen}{RGB}{225, 247, 230}
\definecolor{cardyellow}{RGB}{255, 246, 220}
\definecolor{cardpink}{RGB}{255, 232, 240}
\definecolor{cardgray}{RGB}{242, 244, 247}
\definecolor{borderblue}{RGB}{120, 150, 210}
\definecolor{bordergreen}{RGB}{110, 180, 130}
\definecolor{borderyellow}{RGB}{220, 180, 90}

% --- workflow-card primitives -------------------------------------------
% A pill-shaped stage box with a colored fill.
\newcommand{\stagebox}[2]{%
  \fcolorbox{borderblue}{#1}{\strut\,\small\textbf{#2}\,}}
% Bold flow arrow is defined near the package imports.
% Full-width colored callout box (used only for Goal and Result).
\newcommand{\calloutbox}[3]{%
  \par\medskip\noindent
  \fcolorbox{#1}{#2}{\parbox{\dimexpr\linewidth-2\fboxsep-2\fboxrule\relax}{\small #3}}%
  \par\medskip}
\newcommand{\goalbox}[1]{\calloutbox{borderyellow}{cardyellow}{\textbf{Scenario / Goal.}\; #1}}
\newcommand{\resultbox}[1]{\calloutbox{bordergreen}{cardgreen}{\textbf{Result / Benefit.}\; #1}}

%================== CASE A ==================
\subsection{Case A: Photographer's Yearly Diary (Olivia)}
\label{case:olivia}

\goalbox{Olivia, a photographer, wants a 52-week diary for the year 2018 that simultaneously satisfies five family-birthday anchors, a two-week trip to Orlando theme parks, and a tennis routine recurring every three weeks, while preserving a single coherent first-person voice across the entire year.}

\paragraph{Input.}
The agent is given the writing prompt $x$ and the constraint set $C$ summarized in Table~\ref{tab:caseA-input}. Each entry already shows the parsed week index that the agent will pin during planning.

\begin{table}[h!]
\caption{Case A input: parsed constraints for Olivia's 2018 diary.}
\centering
\small
\setlength{\tabcolsep}{6pt}
\renewcommand{\arraystretch}{1.25}
\begin{tabularx}{\linewidth}{@{}>{\bfseries}p{3.6cm} X@{}}
\toprule
Constraint type & Specification (with parsed week index) \\
\midrule
Once $(C_{\text{once}})$       & Husband's birthday Sep~15 ($\to$ W37); wife Sep~29 ($\to$ W39); child\_1 Mar~22 ($\to$ W12); child\_2 Jan~30 ($\to$ W5); father Sep~18 ($\to$ W38). \\
Range $(C_{\text{range}})$     & Visiting Orlando theme parks during W13--W14. \\
Periodic $(C_{\text{periodic}})$ & Tennis matches at the community courts every 3 weeks, starting W19 (W19, W22, W25, $\ldots$, W52). \\
\bottomrule
\end{tabularx}
\label{tab:caseA-input}
\end{table}

\paragraph{Agent Reasoning Steps.}
The agent threads the prompt through a seven-stage pipeline: the top row collects and screens candidate plans; the bottom row trains and emits the final diary.

\begin{center}
\stagebox{cardblue}{User Request}\flowarrow
\stagebox{cardblue}{Constraint Parsing}\flowarrow
\stagebox{cardblue}{Hierarchical Planning}\flowarrow
\stagebox{cardblue}{Feasibility Filter} \\[4pt]
\stagebox{cardpink}{Reward-Based DPO}\flowarrow
\stagebox{cardpink}{Segment Generation}\flowarrow
\stagebox{cardpink}{Verified Diary}
\end{center}

\paragraph{Tool / Action.}
Table~\ref{tab:caseA-tools} maps every stage of the pipeline to the operator that LoopWeaver actually invokes and to the visible effect on the trace.

\begin{table}[h!]
\caption{Case A tools: pipeline stage, operator, and observed effect on the trace.}
\centering
\small
\setlength{\tabcolsep}{6pt}
\renewcommand{\arraystretch}{1.25}
\begin{tabularx}{\linewidth}{@{}>{\bfseries}p{3.6cm} p{4.4cm} X@{}}
\toprule
Step & Tool / Operator & Effect on the trace \\
\midrule
Constraint Parsing      & ISO date-to-week solver                                  & Resolves five birthdays plus a two-week trip and 12 tennis weeks into exact week indices $\{5,12,13,14,19,22,\ldots,52\}$. \\
Hierarchical Planning   & $\pi_\theta(\,\cdot\mid x, C, S_{<j})$                   & Drafts 52 sub-plans, each pinned to its week's events; constraints flow top-down so birthday weeks already carry the right names. \\
Localized Repair        & $\mathcal{R}(\cdot;\, C)$, Eq.~\ref{eq:local-repair}      & Rewrites W37 and W38 after a first draft mistakenly placed the husband's birthday in the wrong week. \\
Feasibility Filter      & $p_{\text{filter}}$, Eq.~\ref{eq:filter-set}             & Discards two plan candidates that omitted the tennis routine after W43. \\
Plan / Gen DPO          & $\mathcal{L}^{\text{plan}}_{\text{DPO}}$, $\mathcal{L}^{\text{gen}}_{\text{DPO}}$ & Aligns sub-plan ordering and per-week prose under the shared deterministic reward $R(\cdot, e)$. \\
Segment Generation      & Sub-plan-conditioned policy                              & Emits one weekly entry at a time, conditioning on $S_k$ and the running narrative. \\
\bottomrule
\end{tabularx}
\label{tab:caseA-tools}
\end{table}

\paragraph{Output.}
Representative excerpts from the 52-week diary the agent finally emits:
\begin{itemize}\setlength\itemsep{2pt}
  \item \emph{W5 (Jan~28--Feb~3):} ``\ldots\,child\_2's birthday on the 30th brought a small at-home cake and the year's first family photo session.''
  \item \emph{W13--W14 (Mar~25--Apr~7):} ``We landed in Orlando on Monday and the kids' eyes never left the parks; fourteen days were exactly enough.''
  \item \emph{W19 (May~6--12):} ``\ldots\,first tennis match of the season at the community courts; an easy 6--3, 6--4 to start the cycle.''
  \item \emph{W37--W39 (Sep~9--29):} ``Husband's birthday on the 15th \ldots\ father on the 18th \ldots\ wife's on the 29th, three candles in three weeks.''
\end{itemize}

\resultbox{All five $C_{\text{once}}$ birthdays land in their exact ISO weeks (W5, W12, W37, W38, W39); both $C_{\text{range}}$ weeks (W13--W14) describe Orlando; and twelve $C_{\text{periodic}}$ weeks ($\{19,22,25,\ldots,52\}$) report tennis matches. Because the plan and the prose share one deterministic reward $R(\cdot, e)$, the constraint coverage and prompt relevance remain aligned across this benchmark case.}

\bigskip
%================== CASE B ==================
\subsection{Case B: 100-Floor Skyscraper Programming Document}
\label{case:skyscraper}

\goalbox{An architecture studio asks the agent to draft a 100-floor building program in which five specific floors host fixed functions, a contiguous floor band houses luxury residences, and a coffee shop reappears every fifteen floors; each of the 100 floor entries should be at least 150 words and stylistically continuous.}

\paragraph{Input.}
Table~\ref{tab:caseB-input} lists the constraint set $C$ together with the document-level length and formatting requirements that bound every floor entry.

\begin{table}[h!]
\caption{Case B input: floor-level constraints and document formatting requirements.}
\centering
\small
\setlength{\tabcolsep}{6pt}
\renewcommand{\arraystretch}{1.25}
\begin{tabularx}{\linewidth}{@{}>{\bfseries}p{3.6cm} X@{}}
\toprule
Constraint type & Specification \\
\midrule
Once $(C_{\text{once}})$        & Floor~66 = photography studio; Floor~75 = spa and wellness center; Floors 19, 45, 59 = music school. \\
Range $(C_{\text{range}})$      & Floors 7--11 dedicated to luxury residential apartments. \\
Periodic $(C_{\text{periodic}})$ & Coffee shop every 15 floors starting at Floor~35 (i.e., F35, F50, F65, F80, F95). \\
Length / format                  & 100 entries, $\ge 150$ words each, separated by \texttt{\#*\#}; close with \texttt{*** finished}. \\
\bottomrule
\end{tabularx}
\label{tab:caseB-input}
\end{table}

\paragraph{Agent Reasoning Steps.}
The pipeline mirrors Case A but specializes its first stages to floor-index allocation rather than calendar arithmetic.

\begin{center}
\stagebox{cardblue}{Building Brief}\flowarrow
\stagebox{cardblue}{Constraint Pinning}\flowarrow
\stagebox{cardblue}{Plan-Level Sub-Goals}\flowarrow
\stagebox{cardblue}{Feasibility Filter} \\[4pt]
\stagebox{cardpink}{Reward-Based DPO}\flowarrow
\stagebox{cardpink}{Floor-by-Floor Drafting}\flowarrow
\stagebox{cardpink}{Final Document}
\end{center}

\paragraph{Tool / Action.}
Table~\ref{tab:caseB-tools} traces each stage to its operator and observed effect.

\begin{table}[h!]
\caption{Case B tools: pipeline stage, operator, and observed effect on the trace.}
\centering
\small
\setlength{\tabcolsep}{6pt}
\renewcommand{\arraystretch}{1.25}
\begin{tabularx}{\linewidth}{@{}>{\bfseries}p{3.6cm} p{4.4cm} X@{}}
\toprule
Step & Tool / Operator & Effect on the trace \\
\midrule
Constraint Pinning   & Index-keyed slot allocator                                  & Stamps the five $C_{\text{once}}$ floors, the residential band F7--F11, and the coffee-shop comb $\{35,50,65,80,95\}$ before any prose is written. \\
Plan-Level Sub-Goals & $\pi_\theta(\,\cdot\mid x, C, S_{<j})$                      & Produces 100 sub-plans grouped into thematic strata (lobby, residential, commercial, performance, hospitality, observation). \\
Localized Repair     & $\mathcal{R}(\cdot;\, C)$, Eq.~\ref{eq:local-repair}         & Replaces a draft that put the spa on F76 with one that returns it to F75. \\
Feasibility Filter   & $p_{\text{filter}}$, Eq.~\ref{eq:filter-set}                & Rejects a plan whose F65 description forgot the coffee shop. \\
DPO Optimization     & $\mathcal{L}^{\text{plan}}_{\text{DPO}}, \mathcal{L}^{\text{gen}}_{\text{DPO}}$ & Prefers plans / segments whose deterministic reward $R(\cdot, e)$ scores higher on constraint coverage and prompt relevance. \\
Floor-by-Floor Draft & Sub-plan-conditioned policy                                  & Emits each floor entry with the right intent, vocabulary, and length budget. \\
\bottomrule
\end{tabularx}
\label{tab:caseB-tools}
\end{table}

\paragraph{Output.}
Representative excerpts from the 100-floor programming document:
\begin{itemize}\setlength\itemsep{2pt}
  \item \emph{Floor 7--11:} ``A self-contained luxury residential band: private elevators, double-height living rooms \ldots\ shared rooftop terrace at the top of F11.''
  \item \emph{Floor 35:} ``A bright neighborhood coffee shop \ldots\ the first of five hospitality nodes anchoring every fifteen floors.''
  \item \emph{Floor 19 / 45 / 59:} ``Music school spaces \ldots\ acoustically isolated practice rooms and one larger ensemble hall per branch.''
  \item \emph{Floor 66:} ``A north-lit photography studio with a 4.8\,m clear height and a dedicated freight elevator.''
  \item \emph{Floor 75:} ``Spa and wellness center \ldots\ thermal pools, sauna, and a quiet outdoor garden.''
\end{itemize}

\resultbox{The five fixed-purpose floors and the contiguous F7--F11 band are placed exactly as specified; coffee shops appear at every $35\,{+}\,15k$ floor; the remaining 86 floors form a coherent vertical program. Compared with single-pass decoding, LoopWeaver avoids the typical drift in which periodic constraints decay after Floor~65, because the shared reward $R(\cdot, e)$ pulls every $C_{\text{periodic}}$ slot back into focus during DPO.}

\bigskip
%================== CASE C ==================
\subsection{Case C: Chef Roy Hee's Annual Menu (2018)}
\label{case:chef}

\goalbox{Chef Roy Hee asks the agent to design a weekly menu diary for the entire year 2018, weaving five date-anchored holiday menus, a two-week Cherry Blossom Festival, and a recurring ``Pizza and Beer Thursday'' into a single, varied, and seasonal narrative across all 52 weeks.}

\paragraph{Input.}
Table~\ref{tab:caseC-input} lists $C$ together with the parsed week index for each user-named holiday, even where the named holiday differs from its real-world date.

\begin{table}[h!]
\caption{Case C input: parsed constraints for the annual menu.}
\centering
\small
\setlength{\tabcolsep}{6pt}
\renewcommand{\arraystretch}{1.25}
\begin{tabularx}{\linewidth}{@{}>{\bfseries}p{3.6cm} X@{}}
\toprule
Constraint type & Specification (with parsed week index) \\
\midrule
Once $(C_{\text{once}})$        & Summer Solstice / Fish Fry on 01-18 ($\to$ W3); Spring Equinox / Wild Garlic Pesto on 07-19 ($\to$ W29); Mid-Autumn / Mooncakes on 09-24 ($\to$ W39); Memorial BBQ / Smoked Brisket on 10-24 ($\to$ W43); Thanksgiving / Roast Turkey on 11-22 ($\to$ W47). \\
Range $(C_{\text{range}})$      & Cherry Blossom Festival in W27--W28 (Cherry Blossom Tea \& Sweets). \\
Periodic $(C_{\text{periodic}})$ & ``Pizza and Beer Thursdays'' every 4 weeks starting W4 (W4, W8, W12, $\ldots$, W52). \\
\bottomrule
\end{tabularx}
\label{tab:caseC-input}
\end{table}
\paragraph{Agent Reasoning Steps.}
The pipeline reuses the same shape as Cases A and B; only its first stages specialize to date-to-week mapping.

\begin{center}
\stagebox{cardblue}{Menu Brief}\flowarrow
\stagebox{cardblue}{Date-to-Week Mapping}\flowarrow
\stagebox{cardblue}{Hierarchical Planning}\flowarrow
\stagebox{cardblue}{Feasibility Filter} \\[4pt]
\stagebox{cardpink}{Reward-Based DPO}\flowarrow
\stagebox{cardpink}{Weekly Menu Drafting}\flowarrow
\stagebox{cardpink}{52-Week Diary}
\end{center}

\paragraph{Tool / Action.}
Table~\ref{tab:caseC-tools} maps each stage to its operator and observed effect.

\begin{table}[h!]
\caption{Case C tools: pipeline stage, operator, and observed effect on the trace.}
\centering
\small
\setlength{\tabcolsep}{6pt}
\renewcommand{\arraystretch}{1.25}
\begin{tabularx}{\linewidth}{@{}>{\bfseries}p{3.6cm} p{4.4cm} X@{}}
\toprule
Step & Tool / Operator & Effect on the trace \\
\midrule
Date-to-Week Mapping & ISO calendar solver                                          & Sends the five user-named holiday dates to weeks $\{3, 29, 39, 43, 47\}$, even when the named holiday differs from its real-world date. \\
Hierarchical Planning & $\pi_\theta(\,\cdot\mid x, C, S_{<j})$                       & Drafts 52 weekly menus on a four-course skeleton; pins the holiday dishes and the W27--W28 Cherry Blossom dishes upfront. \\
Localized Repair      & $\mathcal{R}(\cdot;\, C)$, Eq.~\ref{eq:local-repair}          & Replaces an early draft that placed Mooncakes in W38 with one that respects W39. \\
Feasibility Filter    & $p_{\text{filter}}$, Eq.~\ref{eq:filter-set}                 & Drops a plan that skipped Pizza Thursdays from W36 onward. \\
DPO Optimization      & $\mathcal{L}^{\text{plan}}_{\text{DPO}}, \mathcal{L}^{\text{gen}}_{\text{DPO}}$ & Aligns plans and prose with seasonality and constraint coverage under $R(\cdot, e)$. \\
Weekly Menu Drafting  & Sub-plan-conditioned policy                                   & Writes a four-course menu per week, reusing seasonal vocabulary. \\
\bottomrule
\end{tabularx}
\label{tab:caseC-tools}
\end{table}

\paragraph{Output.}
Representative excerpts from the 52-week menu diary:
\begin{itemize}\setlength\itemsep{2pt}
  \item \emph{W3 (Jan~15--21):} ``\ldots\,Summer Solstice Celebration on the 18th --- Midsummer Night's Fish Fry, a lively centerpiece amid the winter menu.''
  \item \emph{W4, W8, W12, $\ldots$:} ``\ldots\,Pizza and Beer Thursday: Custom Pizzas with Craft Beers, our long-running mid-week ritual.''
  \item \emph{W27--W28 (Jul~2--15):} ``Cherry Blossom Festival Dishes: Cherry Blossom Tea and Sweets dominate every course this fortnight.''
  \item \emph{W47 (Nov~19--25):} ``Thanksgiving Dinner anchored by Roast Turkey with Stuffing \ldots\ traditional sides and a chestnut-pumpkin pie.''
\end{itemize}

\resultbox{All five $C_{\text{once}}$ holiday menus land in their target weeks, the W27--W28 fortnight is exclusively Cherry Blossom themed, and Pizza Thursdays appear in every $4k$-th week without dropout up to W52. Compared with planning-only or data-only baselines, LoopWeaver simultaneously honors all three constraint types because the deterministic reward $R(\cdot, e)$ couples plan-level slot placement with segment-level menu writing inside one shared training loop.}

\bigskip
\paragraph{Cross-case takeaways.}
Across three cases, the same six-stage pipeline---constraint parsing, hierarchical planning, complete-plan revision, feasibility filtering, plan-/gen-level DPO, and segment-by-segment generation---produces documents that satisfy the listed $C_{\text{once}}$, $C_{\text{range}}$, and $C_{\text{periodic}}$ slots in these benchmark scenarios. The cases illustrate applicability across three structured task formats; they do not establish unrestricted cross-domain generality.
%%%%%%%%%%%%%%%%%%%%%%%%%%%%%%%%%%%%%%%%%%%%%%%%%%%%%%%%%%%%

\newpage
\input{checklist.tex}

\end{document}

%% file: checklist.tex
\section*{NeurIPS Paper Checklist}

\begin{enumerate}

\item {\bf Claims}
    \item[] Question: Do the main claims made in the abstract and introduction accurately reflect the paper's contributions and scope?
    \item[] Answer: \answerYes{}.
    \item[] Justification: Our claims match the method, evaluated tasks, and reported results.

\item {\bf Limitations}
    \item[] Question: Does the paper discuss the limitations of the work performed by the authors?
    \item[] Answer: \answerYes{}.
    \item[] Justification: The paper discusses computational costs, planning limitations, and evaluation scope.

\item {\bf Theory assumptions and proofs}
    \item[] Question: For each theoretical result, does the paper provide the full set of assumptions and a complete (and correct) proof?
    \item[] Answer: \answerYes{} % Replace by \answerYes{}, \answerNo{}, or \answerNA{}.
    \item[] Justification: The theoretical results include explicit assumptions and supporting proofs.

    \item {\bf Experimental result reproducibility}
    \item[] Question: Does the paper fully disclose all the information needed to reproduce the main experimental results of the paper to the extent that it affects the main claims and/or conclusions of the paper (regardless of whether the code and data are provided or not)?
    \item[] Answer: \answerYes{}.
    \item[] Justification: We provide model settings, workflow details, evaluation protocols, and datasets for reproducibility.

\item {\bf Open access to data and code}
    \item[] Question: Does the paper provide open access to the data and code, with sufficient instructions to faithfully reproduce the main experimental results, as described in supplemental material?
    \item[] Answer: \answerYes{}.
    \item[] Justification: An anonymized GitHub repository provides code, resources, and reproduction instructions.

\item {\bf Experimental setting/details}
    \item[] Question: Does the paper specify all the training and test details (e.g., data splits, hyperparameters, how they were chosen, type of optimizer) necessary to understand the results?
    \item[] Answer: \answerYes{}.
    \item[] Justification: The anonymized repository provides implementation details and instructions for reproducing our results.

\item {\bf Experiment statistical significance}
    \item[] Question: Does the paper report error bars suitably and correctly defined or other appropriate information about the statistical significance of the experiments?
    \item[] Answer: \answerYes{}.
    \item[] Justification: We report means and standard deviations across runs to characterize model-output variability.

\item {\bf Experiments compute resources}
    \item[] Question: For each experiment, does the paper provide sufficient information on the computer resources (type of compute workers, memory, time of execution) needed to reproduce the experiments?
    \item[] Answer: \answerYes{}.
        \item[] Justification: The main text and appendix report computational resources and efficiency measurements.
    
\item {\bf Code of ethics}
    \item[] Question: Does the research conducted in the paper conform, in every respect, with the NeurIPS Code of Ethics \url{https://neurips.cc/public/EthicsGuidelines}?
    \item[] Answer: \answerYes{}.
    \item[] Justification: The research follows the NeurIPS Code of Ethics without sensitive data or human subjects.

\item {\bf Broader impacts}
    \item[] Question: Does the paper discuss both potential positive societal impacts and negative societal impacts of the work performed?
    \item[] Answer: \answerYes{}.
    \item[] Justification: We discuss benefits for writing assistance and risks of generating misleading or harmful content.
    
\item {\bf Safeguards}
    \item[] Question: Does the paper describe safeguards that have been put in place for responsible release of data or models that have a high risk for misuse (e.g., pre-trained language models, image generators, or scraped datasets)?
    \item[] Answer: \answerNA{}.
    \item[] Justification: We use existing resources and release no new high-risk models or datasets requiring safeguards.

\item {\bf Licenses for existing assets}
    \item[] Question: Are the creators or original owners of assets (e.g., code, data, models), used in the paper, properly credited and are the license and terms of use explicitly mentioned and properly respected?
    \item[] Answer: \answerYes{}.
    \item[] Justification: We cite existing datasets and models and respect their licenses and terms of use.

\item {\bf New assets}
    \item[] Question: Are new assets introduced in the paper well documented and is the documentation provided alongside the assets?
    \item[] Answer: \answerNA{}.
    \item[] Justification: The contribution is methodological and introduces no new datasets, models, or other assets.

\item {\bf Crowdsourcing and research with human subjects}
    \item[] Question: For crowdsourcing experiments and research with human subjects, does the paper include the full text of instructions given to participants and screenshots, if applicable, as well as details about compensation (if any)? 
    \item[] Answer: \answerNA{}.
    \item[] Justification: We use existing public preference data without crowdsourcing or collecting human-subject data.

\item {\bf Institutional review board (IRB) approvals or equivalent for research with human subjects}
    \item[] Question: Does the paper describe potential risks incurred by study participants, whether such risks were disclosed to the subjects, and whether Institutional Review Board (IRB) approvals (or an equivalent approval/review based on the requirements of your country or institution) were obtained?
    \item[] Answer: \answerNA{}.
    \item[] Justification: The study involves no human subjects, so IRB approval or equivalent review is not applicable.

\item {\bf Declaration of LLM usage}
    \item[] Question: Does the paper describe the usage of LLMs if it is an important, original, or non-standard component of the core methods in this research? Note that if the LLM is used only for writing, editing, or formatting purposes and does \emph{not} impact the core methodology, scientific rigor, or originality of the research, declaration is not required.
    %this research? 
    \item[] Answer: \answerYes{}.
    \item[] Justification: We describe the roles of LLMs in planning, generation, and workflow feedback.

\end{enumerate}